\documentclass[10pt,twocolumn,letterpaper]{article}

\usepackage{iccv}
\usepackage{times}
\usepackage{graphicx}
\usepackage{amsmath}
\usepackage{amssymb}
\usepackage{booktabs}
\usepackage{rotating}
\usepackage{paralist}

\usepackage[linesnumbered,boxed]{algorithm2e}

\usepackage{makecell}

\usepackage{times}

\usepackage{graphicx}
\usepackage{amsmath}
\usepackage{amssymb}
\usepackage[pagebackref=true,breaklinks=true,colorlinks,bookmarks=false,citecolor=blue,linkcolor=blue]{hyperref}
\usepackage{makecell}
\usepackage[accsupp]{axessibility}
\usepackage[table]{xcolor}
\usepackage{color, colortbl}
\definecolor{LightCyan}{rgb}{0.88,1,1}

\hypersetup{
    colorlinks=true,
    urlcolor=magenta,
    citecolor=blue,
}

\makeatletter\if@twocolumn\PassOptionsToPackage{switch}{lineno}\else\fi\makeatother

\usepackage[T1]{fontenc}
\usepackage{multirow}
\usepackage{booktabs}
\usepackage{comment}
\usepackage{color}

\usepackage{textcomp}
\usepackage{siunitx}
\usepackage{tabulary}
\usepackage{makecell}
\usepackage{multirow}
\usepackage{booktabs}
\usepackage{hyperref}
\usepackage{multicol}

\makeatletter
\newcommand\figref{Figure~\ref}
\newcommand{\tabref}[1]{Table~\ref{#1}}
\newcolumntype{P}[1]{>{\centering\arraybackslash}p{#1}}
\newcolumntype{M}[1]{>{\centering\arraybackslash}m{#1}}
\let\ts@includegraphics\includegraphics

\newcommand{\figmargin}{\vspace{-2mm}}

\usepackage{float}

\usepackage[capitalize]{cleveref}
\crefname{section}{Sec.}{Secs.}
\Crefname{section}{Section}{Sections}
\Crefname{table}{Table}{Tables}
\crefname{table}{Tab.}{Tabs.}
\iccvfinalcopy 

\ificcvfinal\fi

\begin{document}


\title{Counting Crowds in Bad Weather}

\author{
Zhi-Kai Huang\textsuperscript{1*}
\quad Wei-Ting Chen\textsuperscript{1,2*}
\quad Yuan-Chun Chiang\textsuperscript{1}
\quad Sy-Yen Kuo\textsuperscript{1} 
\quad Ming-Hsuan Yang\textsuperscript{3,4}
\\\\
\hspace{-8mm}\textsuperscript{1}National Taiwan University\quad \textsuperscript{2}Stanford University
\quad \textsuperscript{3}University of California, Merced
\quad \textsuperscript{4}Google Research
}

\maketitle
\ificcvfinal\thispagestyle{empty}\fi


\let\thefootnote\relax\footnotetext{* indicates equal contribution.}
\begin{abstract}
Crowd counting has recently attracted significant attention in the field of computer vision due to its wide applications to image understanding. 
Numerous methods have been proposed and achieved state-of-the-art performance for real-world tasks. 
However, existing approaches do not perform well under adverse weather such as haze, rain, and snow since the visual appearances of crowds in such scenes are drastically different from those images in clear weather of typical datasets. 
In this paper, we propose a method for robust crowd counting in adverse weather scenarios. 
Instead of using a two-stage approach that involves image restoration and crowd counting modules, 
our model learns effective features and adaptive queries to account for large appearance variations. 
With these weather queries, the proposed model can learn the weather information according to the degradation of the input image and optimize with the crowd counting module simultaneously. 
Experimental results show that the proposed algorithm is effective in counting crowds under different weather types on benchmark datasets.
The source code and trained models will be made available to the public. 
\end{abstract}
\section{Introduction}
\begin{figure}[t!]
\centering \includegraphics[width=0.49\textwidth,page=1]{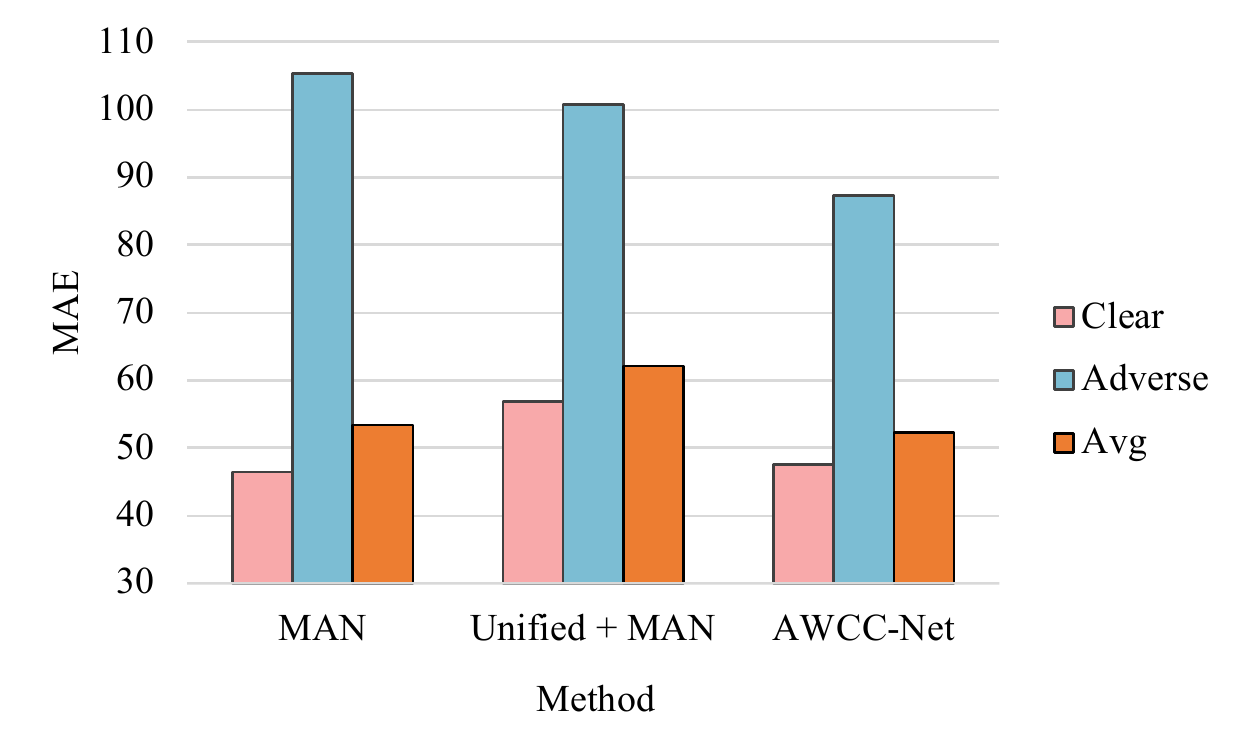}{}
\makeatother 
\caption{\textbf{Performance of State-of-the-art crowd counting methods under adverse and clear weather on the JHU-Crowd++~\cite{sindagi2020jhu} dataset using mean absolute error (MAE).} The MAN method~\cite{lin2022boosting} achieves low MAE in clear scenes but high MAE in adverse weather. 
On the other hand, the two-stage method, based on Unified~\cite{chen2022learning} and MAN~\cite{lin2022boosting}), performs slightly better in adverse weather but slightly worse in clear scenes. 
Overall, the proposed AWCC-Net performs favorably in both scenarios.}
\label{fig:teasor_fig}
\end{figure}

Crowd counting aims to estimate the number of persons in a scene. 
With the advances of deep learning and  construction of large-scale datasets~\cite{zhang2016single,idrees2018composition,sindagi2020jhu,wang2020nwpu}, this topic has become an important topic with numerous applications.
State-of-the-art approaches are mainly based on convolutional neural networks (CNNs)~\cite{jiang2020attention,li2018csrnet,liu2022leveraging,wang2021self,wan2021generalized,gao2019domain,abousamra2021localization,liang2022focal,lin2022boosting} and transformers ~\cite{tian2021cctrans,sun2021boosting,qian2022segmentation,liang2022end}. 
While these approaches count crowds effectively on normal images, they typically do not perform well under adverse weather conditions such as haze, rain, and snow. 
However, adverse weather is a common and inevitable scenario that causes large appearance variations of crowd scenes, thereby significantly affecting the performance of methods developed for clear weather.  
As shown in~\figref{fig:teasor_fig}, the state-of-the-art crowd counting method~\cite{lin2022boosting} performs well only under good weather conditions.
Thus, it is of great importance to develop robust crowd counting methods for adverse weather conditions. 

To solve the crowd counting problem under adverse weather, a plausible solution is to consider a two-stage model. 
Specifically, this strategy pre-processes images using the state-of-the-art image restoration modules~\cite{chen2022learning,valanarasu2022transweather} and then applies the state-of-the-art crowd counting method~\cite{lin2022boosting}. 
However, this two-stage method may not perform well due to several factors.  
First, adopting existing image restoration methods does not always facilitate the crowd counting task significantly since these methods are designed to restore image contents rather than visual classification or regression.
As shown in~\figref{fig:teasor_fig}, a two-stage approach does not address this problem effectively. 
Second, this strategy requires to collect and label images under adverse weather conditions for the restoration process. 
In addition, a two-stage strategy may increase computational complexity significantly. 

In this work, we propose a method based on a transformer to robustly count crowds under adverse and clear scenes without exploiting typical image restoration modules. 
We develop the \textbf{a}dverse \textbf{w}eather \textbf{c}rowd \textbf{c}ounting network (AWCC-Net) which leverages learned weather queries for robust crowd counting. 
In our model, the weather queries are adopted to match the keys and values extracted from the VGG encoder~\cite{simonyan2014very} based on the cross-attention design.
However, the learned weather queries cannot well represent the weather information without appropriate constraints. 
To improve this situation, we present a contrastive weather-adaptive module to improve the learned weather queries. 
This adaptive module computes a weight vector which is combined with the learned weather bank to construct input-dependent weather queries. 
We use the proposed contrastive loss to enforce the learned weather queries to be weather-related. 
With this weather-adaptive module, more robust weather queries can be learned which help the network understand and adapt to the various weather degradation types in the input image.

Extensive experimental results show that the proposed AWCC-Net model performs robustly and favorably against the state-of-the-art schemes for crowd counting under adverse weather conditions without weather annotations. 
We make the following contributions in this work:
\begin{compactitem}
\item We present the AWCC-Net for crowd counting under adverse weather. Our method is based on a transformer architecture that can learn weather-aware crowd counting.
\item We propose a module to constrain the learned queries to be weather-relevant, thereby improving model robustness under adverse weather.
\end{compactitem}

 \section{Related Work}
\subsection{Crowd Counting}
Existing crowd counting methods are mainly based on: (i) detection~\cite{liu2018decidenet,sam2020locate,liu2019point,wang2021self}, (ii) density map~\cite{gao2019domain,idrees2018composition,abousamra2021localization,gao2020learning,liang2022focal,xu2022autoscale,jiang2020attention, sindagi2017generating}, and (iii) regression~\cite{jiang2020attention, sindagi2017generating,idrees2013multi,chen2012feature,chan2009bayesian,song2021rethinking}.
\noindent \smallskip\\
\textbf{Counting by detection.} These approaches use detectors, e.g., Faster RCNN~\cite{ren2015faster}, for crowd counting.
 In~\cite{liu2019point}, a method based on curriculum learning detects and learns
 to predict and count bounding boxes of persons via a locally-constrained regression loss.
 Sam~\etal\cite{sam2020locate} leverages the nearest neighbor distance to generate pseudo bounding boxes and adopt the winner-take-all loss to optimize the box selection during the training stage.
 This technique can benefit the optimization of images with higher resolutions.
\noindent \smallskip\\
\textbf{Counting by density map.} In this mainstream approach, the crowd count is derived via summing over the estimated density map of a scene.
In~\cite{idrees2018composition}, the Gaussian Kernel is applied to construct the density map and propose the composition loss to optimize the crowd counting. 
The DACC method~\cite{gao2019domain} leverages inter-domain features segregation to generate coarse counter and then applies Gaussian prior to compute the counting results.
Abousamra~\etal\cite{abousamra2021localization} develops the topological constraint, which is achieved by a persistence loss to improve the spatial arrangement of dots for density map-based methods. 
Recently, the focal inverse distance transform map~\cite{liang2022focal} shows better performance in representing person locations compared to the aforementioned Gaussian-based density map.
\noindent
\smallskip\\
\textbf{Counting by regression.} The number of the crowd can be regressed from the contextual information among extracted features of cropped image patches.
In~\cite{chan2009bayesian}, a closed-form approximation based on the Bayesian Poisson regression is proposed to compute the crowd counting.
The Fourier analysis and SIFT features can be used to estimate the number of persons in a crowded scene~\cite{idrees2013multi}. 
 On the other hand, the surrogate regression~\cite{song2021rethinking} achieves crowd counting and localization based on a set of point proposals.

Although the aforementioned methods can solve the crowd counting problem on clear images effectively, they do not perform well when the input images are taken in adverse weather. 
Thus, it is of great importance to develop a solution to cope with this problem.
\begin{figure*}[t]
\centering \includegraphics[width=.9\textwidth, page=1]{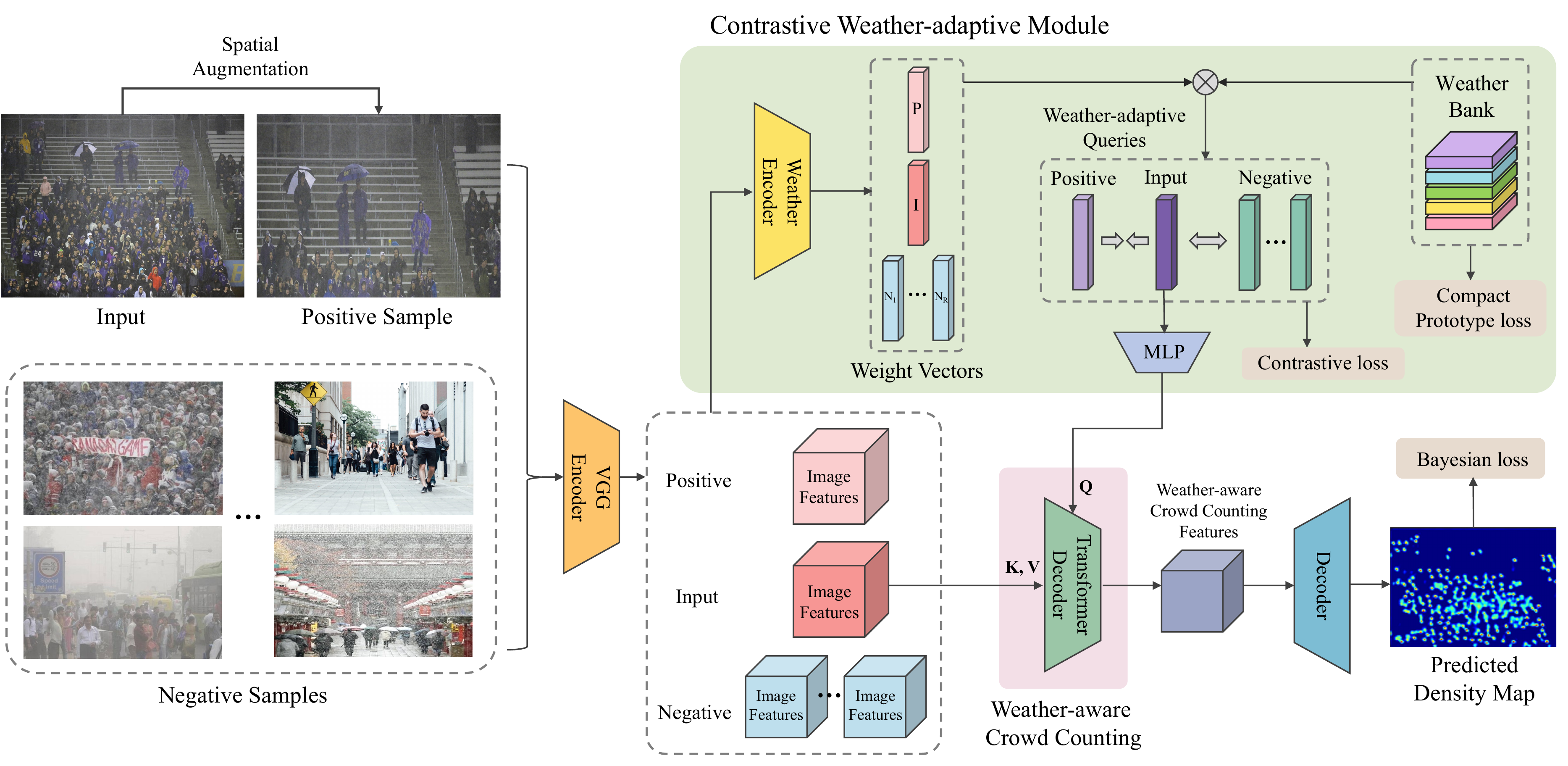}{}
\makeatother 
\caption{\textbf{AWCC-Net for adverse weather crowd counting.} A VGG encoder is adopted to extract image features, which are fed into the contrastive weather-adaptive module to generate patch-wise weather queries. 
Then, the transformer decoder leverages crowd counting features and weather-adaptive queries to generate weather-aware crowd counting features. The decoder computes the density map based on the weather-aware crowd counting features.}
\label{fig:architecture}
\end{figure*}

\subsection{Image Restoration}
In recent years, numerous restoration algorithms have been proposed to handle images acquired in adverse weather, which can be categorized as: (i) single-weather removal; (ii) multi-degradation removal; and (iii) all-in-one weather removal.
\noindent \smallskip\\
\textbf{Single-purpose image restoration.} These methods are developed to restore image contents degraded by one specific degradation such as rain~\cite{jiang2017novel,fu2017removing,yang2019joint,yang2019scale,wei2017should,wang2019spatial,jiang2020multi,du2020variational,wang2020model,wang2021rain,ye2021closing,yang2020wavelet}, snow~\cite{zheng2013single,620844:14515250,620844:14582690,620844:14582739,li2019single,zhang2021deep,chen2021all}, and haze~\cite{he2010single,berman2016non,zhang2018densely,hong2020distilling,deng2020hardgan,shao2020domain,qin2020ffa,chen2019pms}. While these methods are effective for specific conditions, they do not generate clear images when inputs are degraded by other or multiple weather types.
\noindent \smallskip\\
\textbf{Multi-purpose image restoration.} These methods aim to recover various weather types using a unified model. 
A general architecture is developed by Pan~\etal~\cite{pan2018learning} to estimate structures and details simultaneously in parallel branches. 
In~\cite{zamir2021multi}, the MPRNet exploits a multi-stage strategy and  an attention module to refine the incoming features at each stage for effective image restoration. 
Although it is based on a unified framework, different model weights need to be learned for each weather condition. 
\noindent \smallskip\\
\textbf{All-purpose image restoration.} In recent years, much effort has been made to all-purpose image restoration since it only requires a set of pre-trained weights to recover images degraded by different factors. 
An end-to-end training scheme based on the neural architecture search is proposed by Li~\etal~\cite{li2020all} to investigate crucial features from multiple encoders for different weather types and then reconstruct the clear results.
In~\cite{chen2022learning}, a two-stage learning strategy and the multi-contrastive regularization based on knowledge distillation~\cite{hinton2015distilling} are developed to recover multi-weather types. 
Most recently, the TransWeather model~\cite{valanarasu2022transweather} leverages the intra-patch visual information to extract fine detail features for the all-purpose weather removal.

Although these methods can be adopted in pre-process stage for crowd counting under adverse weather, they are designed to recover image content (visual appearance) rather than estimate head counts (visual classification or regression).
Thus, adopting them with crowd counting may have limited performance gain under adverse weather.

\section{Proposed Method}
In this work, we tackle the crowd counting problem under adverse weather.
As shown in \figref{fig:architecture}, image features are first extracted using a pre-trained backbone model, e.g., VGG-19~\cite{simonyan2014very}.
The contrastive weather-adaptive module uses the image features as input to generate the weather-adaptive queries, which are leveraged in the weather-aware crowd counting (WACC) model to generate the weather-aware crowd counting features. 
These features are then fed into the decoder to compute the density map for crowd estimation. 
We discuss the details of these modules in the following sections.

\subsection{Weather-Aware Crowd Counting Model}
\label{sec:WACC}
In this work, we learn weather conditions as queries in our crowd counting model based on a transformer.  
Image features are first extracted via a pre-trained backbone model.
The decoder is similar to that of a vision transformer~\cite{carion2020end}, 
as shown in the pink region of \figref{fig:architecture}. 
The key (K) and value (V) are computed from the image features and the query (Q) is trainable weather queries learned with the network simultaneously. 
Through this operation, the output features of transformer decoder contain crowd counting features with weather information closest to the input scene. 
These features are then fed into a decoder which contains three $3\times3$ convolution blocks with ReLU to predict the density map $D$. 

However, as shown in~\tabref{tab:ablation}, this strategy may obtain limited performance gain since the weather queries are not adaptive to the weather types of the input image as no constraints are enforced. 
That is, the weather information is not learned by the weather queries, and thus they cannot well represent the weather type.
Without effective weather queries, a transformer cannot generate effective features for crowd counting. 

\begin{figure}[t!]
\centering \includegraphics[width=0.48\textwidth,page=1]{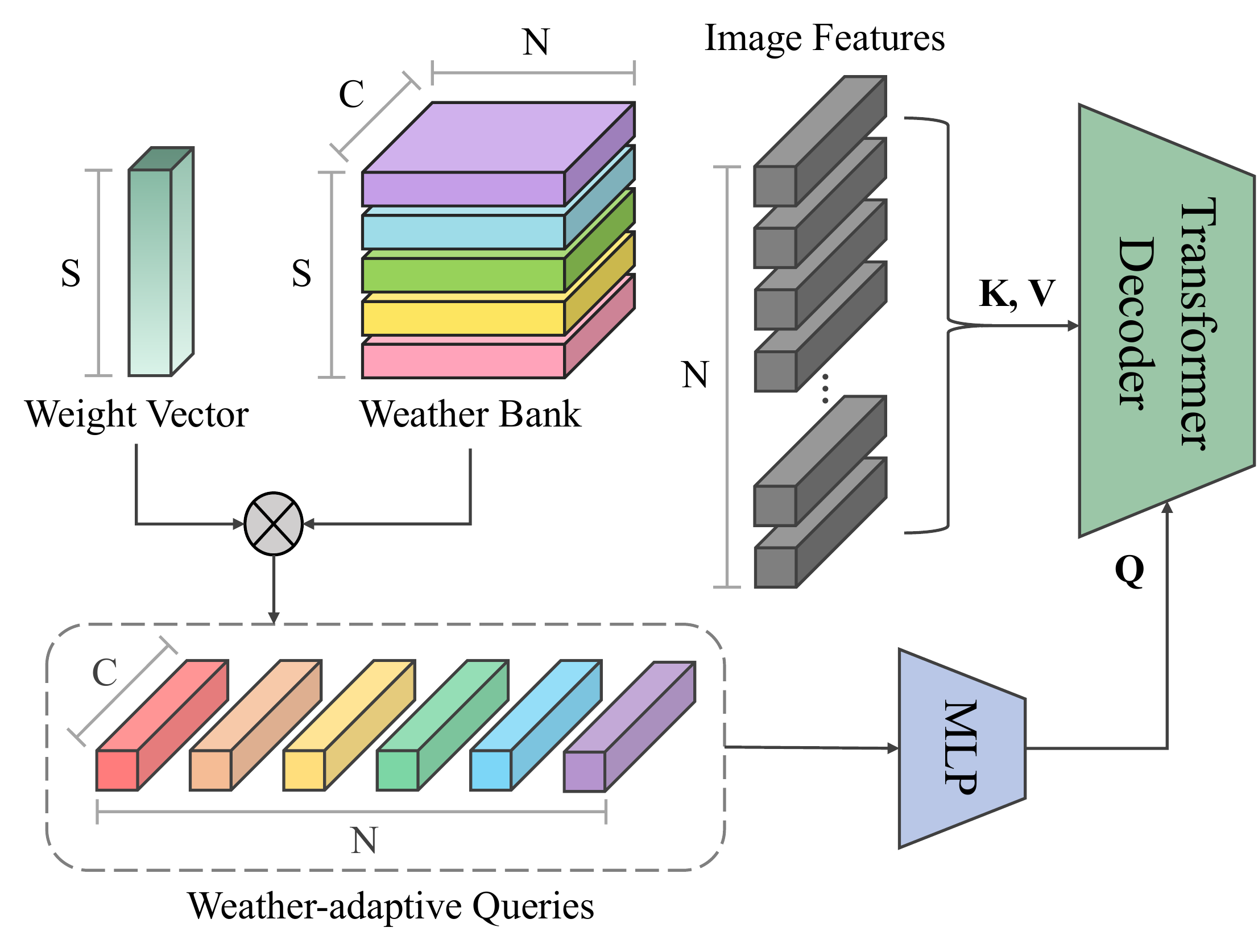}{}
\makeatother 
\caption{\textbf{Learning weather-adaptive queries.} A weight vector is combined with the prototypes in the weather bank via weighted sum to generate $N$ patch-aware weather queries for the transformer decoder.}
\label{fig:WBR}
\end{figure}

\subsection{Contrastive Weather-adaptive Module}
To enforce the learned queries are adaptive to weather conditions, we propose to construct input-dependent weather queries by computing a weight vector. 
The input-dependent weather queries are constrained by the contrastive loss which can guarantee the learned queries to be weather-related. 
\noindent \smallskip\\
\textbf{Pipeline.} The learning process for weather-adaptive queries is shown in the green regions of \figref{fig:architecture} and \figref{fig:WBR}. 
First, the weather encoder predicts a weight vector for the weather bank based on the image features. 
This weight vector can construct the weather queries by the weighted summation of each prototype in the weather bank. 
To constrain these weather queries, we apply the contrastive loss to enforce the weather queries to learn weather information.
A multilayer perceptron (MLP) is then adopted to conduct non-linear operations for the generated weather queries. 
Finally, these weather queries and image features are adopted in the transformer decoder to decode the weather-aware crowd counting features.
\noindent \smallskip\\
\textbf{Weather Bank.} The weather bank stores multiple weather prototypes which can be leveraged to construct weather queries. The dimension of the weather bank is $S \times N \times C$ where $S$, $N$, and $C$ denote the number of weather prototypes, the number of tokens to represent the weather prototype, and the number of channels, respectively. We set $S$, $N$, and $C$ as 8, 48, and 512 in our method. 
These weather prototypes are trainable parameters that are learned with the network simultaneously.
Since they can be learned automatically, our network does not need to adopt the annotations of the weather type to optimize them.
\noindent \smallskip\\
\textbf{Weight Vector.} To construct input-dependent weather queries, we propose to compute a weight vector to represent the weather type of an input image. 
The weight vector is computed by the weather encoder based on the image features whose dimension is $S$. 
With the learned weather bank, the vector can represent the input weather type and construct input-dependent weather queries with dimension $N \times C$ for the contrastive loss.
\noindent \smallskip\\
\textbf{Contrastive Loss.} To learn the weather-adaptive queries, we propose the constrastive loss $\mathcal{L}_{Con}$:
\begin{equation}
\resizebox{1\hsize}{!}{
$\mathcal{L}_{Con}=-\log\left[\frac{\exp(\varphi({v},{{v^{+}}})/\tau)}{\exp(\varphi({v}\cdot{{v^{+}}})/\tau)+\sum_{r=1}^{R}\exp(\varphi({v}\cdot{{v^{-}_{r}}})/\tau)}\right],$}
\label{eq:wc_loss}
\end{equation}
where $v$, $v^{+}$, and $v^{-}$ are the weather queries of the input image, those of the positive sample, and those of the negative samples, respectively.
In addition, $\varphi({\cdot}, \cdot)$ is the cosine similarity function, $\tau$ is the scale temperature, and $R$ denotes the total number of negative samples.

The main idea of our method is that the weather queries of the input image should be similar to those of the positive sample while dissimilar to those of the negative samples. 
To this end, we adopt the random crop and the random flip operations to construct the positive sample. 
Since the positive sample is generated with spatial augmentation, the weather information should be the same as the input image, and thus the two weather queries should be similar. 
We use the rest of the images as our negative samples whose weather queries should be dissimilar.
Although the negative samples may possibly have a similar weather type as the input image, according to~\cite{wang2021understanding}, the contrastive loss contains the tolerance to potential positive samples. 
Thus, the images with the same weather types still have the smaller distance in the feature space. 
We show one example in~\figref{fig:anchor_query}.
\noindent \smallskip\\
\textbf{Compact Prototype Loss.} To reduce the redundancy of learned prototypes in the weather bank, we propose the compact prototype loss $\mathcal{L}_{CP}$:
\begin{equation} 
\mathcal{L}_{CP}=\sum_{i=1}^{S}\sum_{j\neq i}\vert\varphi(P^{i}, P^{j})\vert,
\end{equation}
where $P^{i}$ and $P^{j}$ present the $i^{th}$ and $j^{th}$ prototype in the weather bank. 
With this operation, the prototypes are enforced to be more compact. 

\subsection{Overall Loss}
The proposed AWCC-Net is optimized via multiple loss functions including the compact prototype loss, the contrastive loss, and the crowd counting loss. 
The first two loss functions have been defined in the previous sections. 
The crowd counting loss $\mathcal{L}_{CC}$ aims to constrain the learning of crowd counting and we adopt the Baysesian loss~\cite{ma2019bayesian,lin2022boosting} for the robustness and the better performance:
\begin{equation}
\label{eq:ialoss}
   \mathcal{L}_{CC}=|\sum_{k} \  P_0(k) \cdot D_{k}| + \sum_{i=1}^{L}\ |1 - \sum_{k} \  P_i(k) \cdot D_{k}|,
\end{equation}
where $i$ and $L$ denote the index of the annotated point and the number of annotated points, respectively.
In addition, $D$ is the computed density map, $P_i(k)$ indicates the posterior of the occurrence of the $i^{th}$ annotation given the position $k$, and $P_0(k)$ denotes the background likelihood at position $k$.

The overall loss function of the AWCC-Net is:
\begin{equation}
\mathcal{L}_{AWCC}=\mathcal{L}_{CC}+\lambda_{1}\mathcal{L}_{CP}+\lambda_{2}\mathcal{L}_{Con},
\label{eq:loss_overall}
\end{equation}
where $\lambda_{1}$ and $\lambda_{2}$ are scaling factors.

\section{Implementation Details}
\subsection{Datasets}
We use the ShanghaiTech~\cite{zhang2016single}, UCF-QNRF~\cite{idrees2018composition}, JHU-Crowd++~\cite{sindagi2020jhu}, and NWPU-CROWD~\cite{wang2020nwpu} datasets to evaluate the performance of the proposed method against the state-of-the-art approaches. 
\noindent \smallskip\\
\textbf{JHU-Crowd++}~\cite{sindagi2020jhu}\textbf{.} It consists of 4,372 images and 1.51 million annotated points totally. 
This dataset divides 2,272 images for training, 500 for validation, and the remaining 1,600 images for testing. 
In the testing set, there are 191 images are under adverse weather and 1409 images under normal scenes.
\noindent \smallskip\\
\textbf{ShanghaiTech A}~\cite{zhang2016single}\textbf{.} It contains 482 images and 244, 167 annotated points. 300 images are split for training and the rest of 182 images are for testing.
\noindent \smallskip\\
\textbf{UCF-QNRF}~\cite{idrees2018composition}\textbf{.} It includes 1,535 high-resolution images collected from the Web, with 1.25 million annotated points. 
There are 1,201 images in the training set and 334 images in the testing set. 
The UCF-QNRF dataset has a wide range of people count between 49 and 12,865.
\noindent \smallskip\\
\textbf{NWPU-CROWD}~\cite{wang2020nwpu}\textbf{.} It consists of 5,109 images with 2.13 million annotated points. 3,109 images are divided into the training set and 500 images are in the validation set, and the remaining 1,500 images are for testing.

\subsection{Evaluation Protocol}
Similar to the existing approaches, we apply Mean Absolute Error (MAE) and Mean Squared Error (MSE) for performance evaluation: 
{\small
\begin{equation}
MAE  =\frac{1}{Q}\ \sum_{i=1}^{Q}\big|GT_{i}-N_i\big|, 
MSE =\sqrt{\frac{1}{Q}\ \sum_{i=1}^{Q}(GT_i-N_i)^2},
\end{equation}
}
where $Q$ is the number of images, $GT_i$ and $N_i$ denote the ground truth and  predicted crowd count of the $i$-th image. 

\subsection{Training Details}
In this work, the learning rate is $10^{-5}$, and the Adam optimizer is applied. 
We set the batch size to be 1, and randomly crop all input images to $512\times512$ in the training process.
The proposed network is trained on an Nvidia Tesla V100 GPU and implemented using the PyTorch framework. 
Our model is based on a hybrid CNN-Transformer backbone which contains a VGG-19 model pre-trained on ImageNet, a transformer decoder, and a CNN-based regression layer. 
The scaling factors of the losses $\lambda_1$ and $\lambda_2$ are both set to 1. 
In addition, the total number of negative samples $R$ is 64, and the scale temperature $\tau$ is 0.2.

In the spatial augmentation, we crop two patches of the training image. 
We apply random horizontal flipping on the first patch. 
Then, we crop a second patch overlapped with the first one with an overlap factor sampled from a uniform distribution. 
The first patch serves as the training input for crowd counting and the anchor for the contrastive loss. 
The second patch is the positive sample for the contrastive loss and is not directly involved in crowd counting.

\section{Experimental Results}

In this section, we present the evaluation results of the proposed method for crowd counting in both adverse weather scenes and normal scenes. 
More results are available in the supplementary material.

\begin{table}[t!]
\small
\begin{center}
\scalebox{0.83}{
\begin{tabular}{ccccccccc} 
\toprule
\multirow{2}{*}{\textbf{Method}} & \multicolumn{2}{c}{\textbf{Clear}} & \multicolumn{2}{c}{\textbf{Adverse Weather }}& \multicolumn{2}{c}{\textbf{Average}} \\ 
\cline{2-7}
& MAE & MSE & MAE & MSE & MAE & MSE   \\ 
\hline\hline
SFCN~\cite{Wang2019LearningFS} & 71.4 & 225.3 & 122.8 & 606.3 & 77.5 & 297.6 \\
BL~\cite{ma2019bayesian} & 66.2 & 200.6 & 140.1 & 675.7& 75.0 & 299.9 \\
BL~\cite{ma2019bayesian}-U & 65.3 & 208.4 & 134.0 & 645.6& 73.5 & 296.6 \\
BL~\cite{ma2019bayesian}-UF & 62.6 & 205.7 & 130.4 & 638.4 & 70.7 & 293.1 \\
LSCCNN~\cite{sam2020locate} & 103.8 & 399.2 & 178.0 & 744.3 & 112.7 & 454.4 \\
CG-DRCN-V~\cite{sindagi2020jhu} & 74.7 & 253.4 & 138.6 & 654.0 & 82.3 & 328.0 \\
CG-DRCN-R~\cite{sindagi2020jhu} & 64.4 & 205.9 & 120.0 & 580.8 & 71.0 & 278.6 \\
UOT~\cite{ma2021learning} & 53.1 & 148.2 & 114.9 & 610.7 & 60.5 & 252.7 \\
GL~\cite{wan2021generalized} & 54.2 & 159.8 & 115.9 & 602.1 & 61.6 & 256.5 \\
GL~\cite{wan2021generalized}-U & 64.4 & 207.0 & 127.2 & 617.3 & 71.9 & 288.5 \\
GL~\cite{wan2021generalized}-UF & 56.3 & 174.1 & 127.6 & 658.5 & 64.8 & 280.1 \\
CLTR~\cite{liang2022end} & 52.7 & \underline{148.1} & 109.5 & 568.5 & 59.5 & 240.6 \\
MAN~\cite{lin2022boosting} & \textbf{46.5} & \textbf{137.9} & 105.3 & \underline{478.4} & \underline{53.4} & \underline{209.9} \\
MAN~\cite{lin2022boosting}-U & 56.9 & 182.5 & \underline{100.7} & 548.2 & 62.1 & 255.4 \\
MAN~\cite{lin2022boosting}-UF & 60.8 & 187.7 & 117.1 & 623.2 & 67.6 & 278.2 \\
\rowcolor{LightCyan}
\textbf{AWCC-Net} & \underline{47.6} & 153.9 & \textbf{87.3} & \textbf{430.1}& \textbf{52.3} & \textbf{207.2}\\
\bottomrule
\end{tabular}}
	\end{center}
\caption{\textbf{Quantitative comparison on the JHU-Crowd++~\cite{sindagi2020jhu} dataset with existing methods.} We evaluate the performance in adverse weather scenes and clear scenes. The words with \textbf{boldface} indicate the best results, and those with \underline{underline} indicate the second-best results.}
\label{tab:performance}
\end{table}

\begin{table*}[htb]
\centering
\small
\scalebox{0.95}{
\centering
\begin{tabular}{ccccccccc} 
\toprule
\multicolumn{1}{c}{\textbf{Dataset}} & \multicolumn{2}{c}{\textbf{ShanghaiTechA}} & \multicolumn{2}{c}{\textbf{UCF-QNRF}}  & \multicolumn{2}{c}{\textbf{JHU-Crowd++}}& \multicolumn{2}{c}{\textbf{NWPU-CROWD}}  \\ 
\hline
\multicolumn{1}{c}{\textbf{Method}}  & MAE & \multicolumn{1}{c}{MSE}     & MAE & \multicolumn{1}{c}{MSE} & MAE & MSE & MAE & MSE                 \\ 
\hline\hline
SFCN~\cite{Wang2019LearningFS} & 64.8 & 107.5 & 102.0 & 171.4 & 77.5 & 297.6 & 105.7 & 424.1 \\
BL~\cite{ma2019bayesian} & 62.8 & 101.8 & 88.7 & 154.8 & 75.0 & 299.9 & 105.4 & 454.2 \\
LSCCNN~\cite{sam2020locate} & 66.5 & 101.8 & 120.5 & 218.2 & 112.7 & 454.4 & - & - \\ 
CG-DRCN-VGG16~\cite{sindagi2020jhu} & 64.0 & 98.4 & 112.2 & 176.3 & 82.3 & 328.0& - & - \\
CG-DRCN-Res101~\cite{sindagi2020jhu} & 60.2 & 94.0 & 95.5 & 164.3 & 71.0 & 278.6 & - & - \\
UOT~\cite{ma2021learning} & 58.1 & 95.9 & 83.3 & 142.3 & 60.5 & 252.7 & 87.8 & 387.5 \\
S3~\cite{lin2021direct} & 57.0 & 96.0 & 80.6 & 139.8 & 59.4 & 244.0 & 83.5 & 346.9 \\
GL~\cite{wan2021generalized} & 61.3 & 95.4 & 84.3 & 147.5 & 59.9 & 259.5 & 79.3 & 346.1 \\
ChfL~\cite{shu2022crowd} & 57.5 & 94.3 & 80.3 & 137.6 & 57.0 & 235.7 & 76.8 & 343.0 \\
CLTR~\cite{liang2022end} & 56.9 & 95.2 & 85.8 & 141.3 & 59.5 & 240.6 & \textbf{74.3} & 333.8 \\
MAN~\cite{lin2022boosting}  & 56.8 & \underline{90.3} & \underline{77.3} & \underline{131.5} & \underline{53.4} & \underline{209.9} & 76.5 & \textbf{323.0} \\
GauNet~\cite{cheng2022rethinking} & \textbf{54.8} & \textbf{89.1} & 81.6 & 153.7 & 58.2 & 245.1 & - & - \\
\rowcolor{LightCyan}
\textbf{AWCC-Net} & \underline{56.2} & 91.3 & \textbf{76.4} & \textbf{130.5} & \textbf{52.3} & \textbf{207.2}& \underline{74.4} & \underline{329.1} \\
\bottomrule
\end{tabular}
 }
 \vspace{1mm}
\caption{\textbf{Quantitative comparison on the ShanghaiTech A~\cite{zhang2016single}, UCF-QNRF~\cite{idrees2018composition}, JHU-Crowd++~\cite{sindagi2020jhu}, and NWPU-CROWD~\cite{wang2020nwpu} datasets with existing methods.} The words with \textbf{boldface} indicate the best results, and those with \underline{underline} indicate the second-best results.}
\label{tab:performance_other}
\end{table*}

\begin{figure*}[t!]
\begin{center}
	\footnotesize
    \hspace{0.006\textwidth}
     \begin{minipage}[c]{0.195\textwidth}
      \centering    
        \text{Input}
    \end{minipage}
    \begin{minipage}[c]{0.195\textwidth}
     \centering
        \text{Ground Truth}
    \end{minipage}
         \begin{minipage}[c]{0.195\textwidth}
     \centering
        \text{MAN~\cite{lin2022boosting}}   
    \end{minipage}
         \begin{minipage}[c]{0.195\textwidth}
     \centering
        \text{Unified~\cite{chen2022learning}+MAN~\cite{lin2022boosting}}    
    \end{minipage}
         \begin{minipage}[c]{0.195\textwidth}
     \centering
        \text{AWCC-Net}
    \end{minipage}
    \\
    \begin{minipage}[c]{0.02\textwidth}
    \rotatebox[origin = c]{90}{Haze}
    \end{minipage}
    \begin{minipage}[c]{0.975\textwidth}
    \centering
        \includegraphics[width=0.195\textwidth]{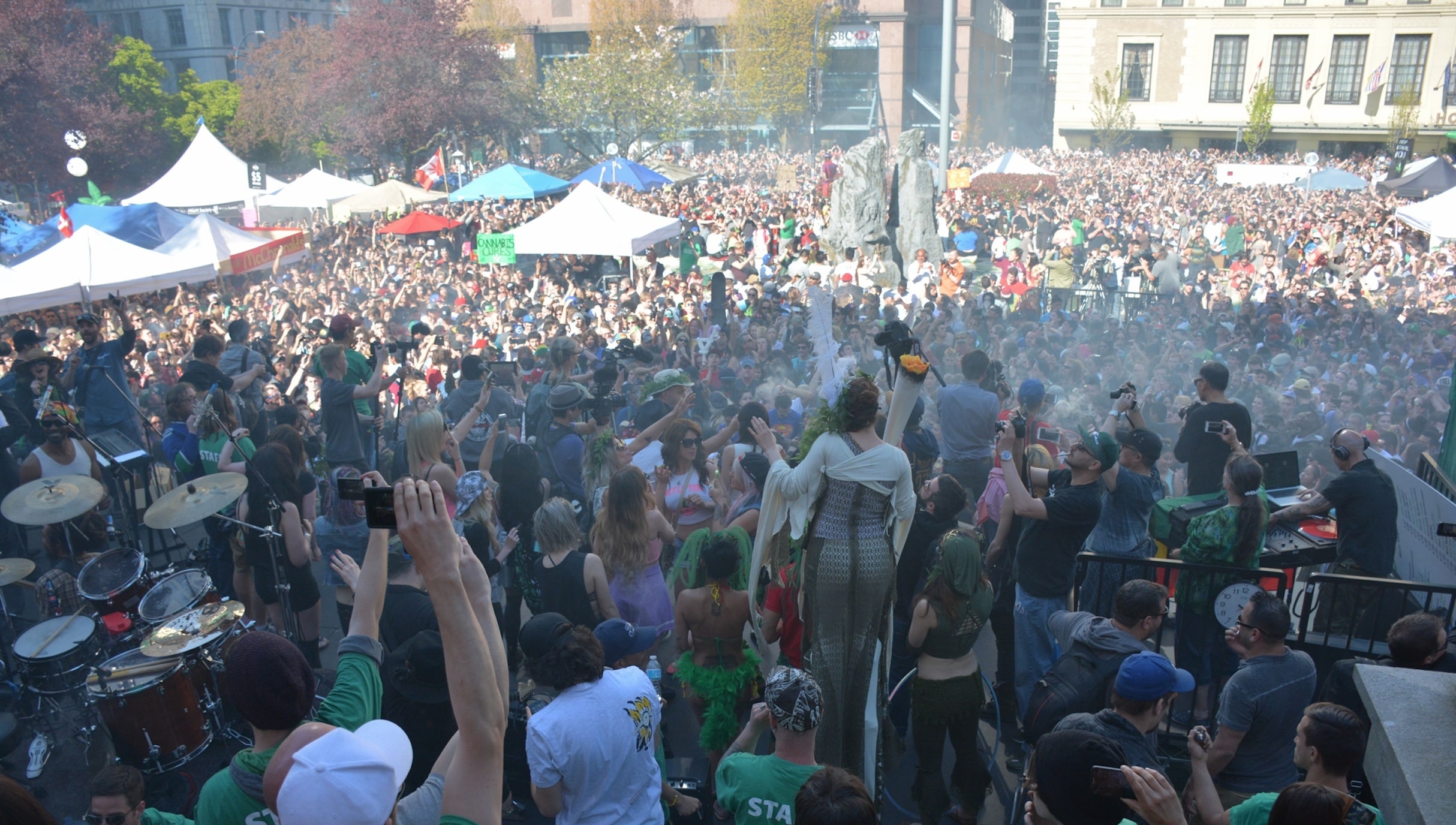} \hfill
        \includegraphics[width=0.195\textwidth]{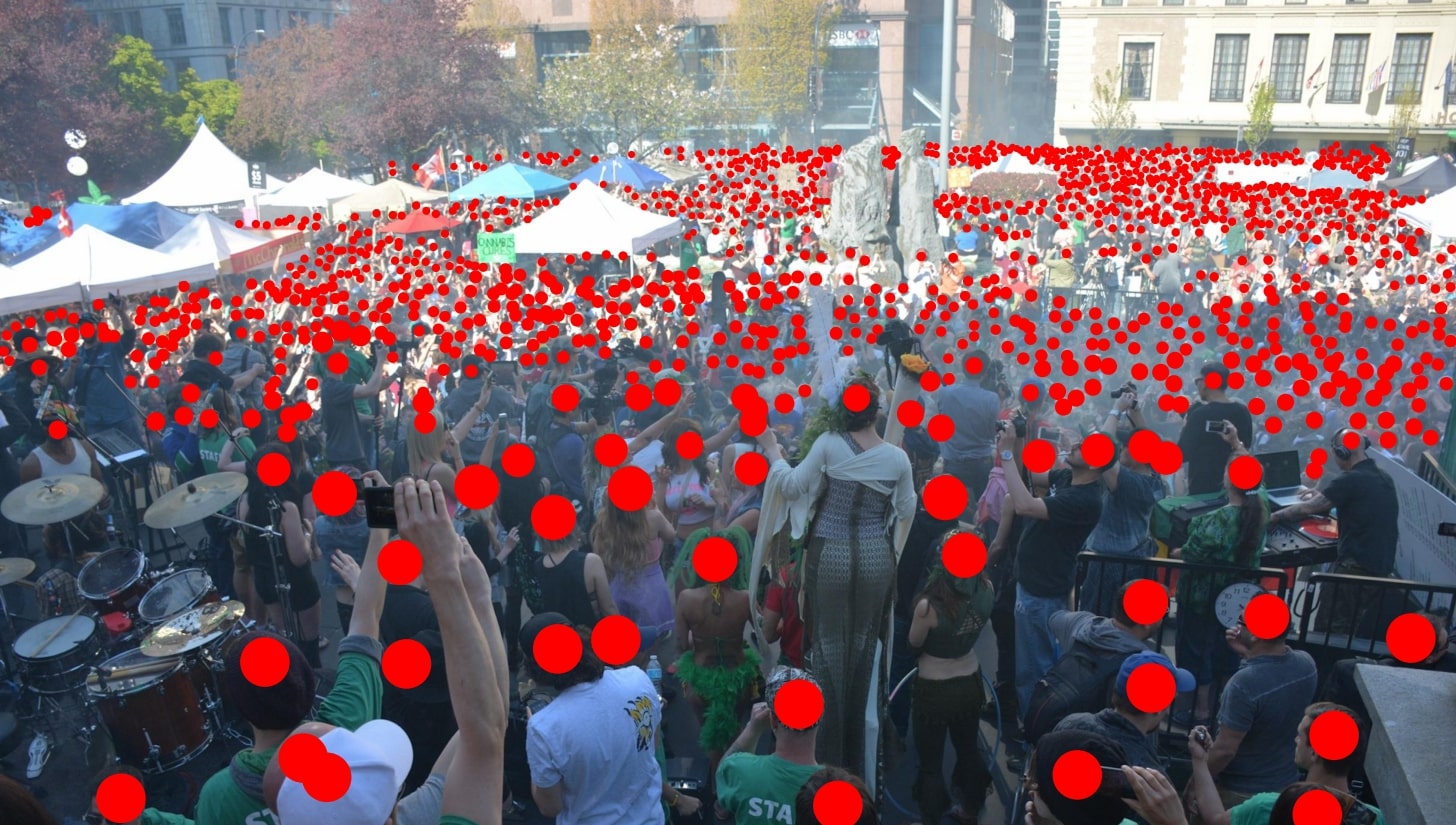} \hfill
        \includegraphics[width=0.195\textwidth]{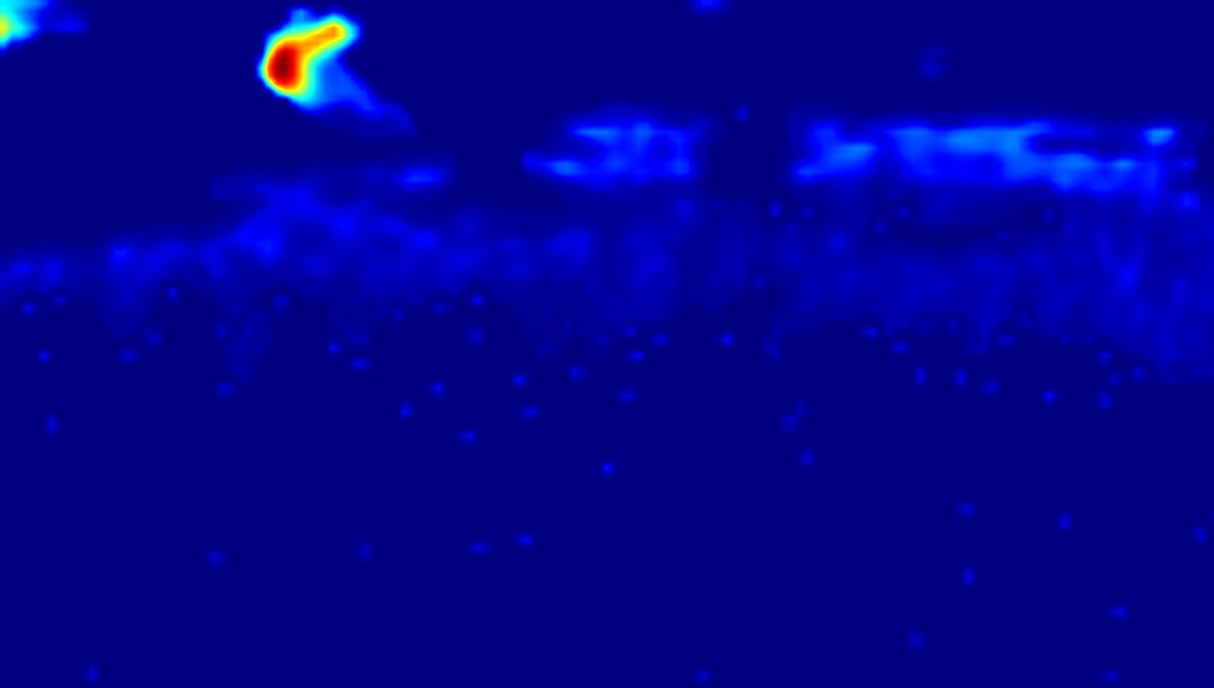} \hfill
        \includegraphics[width=0.195\textwidth]{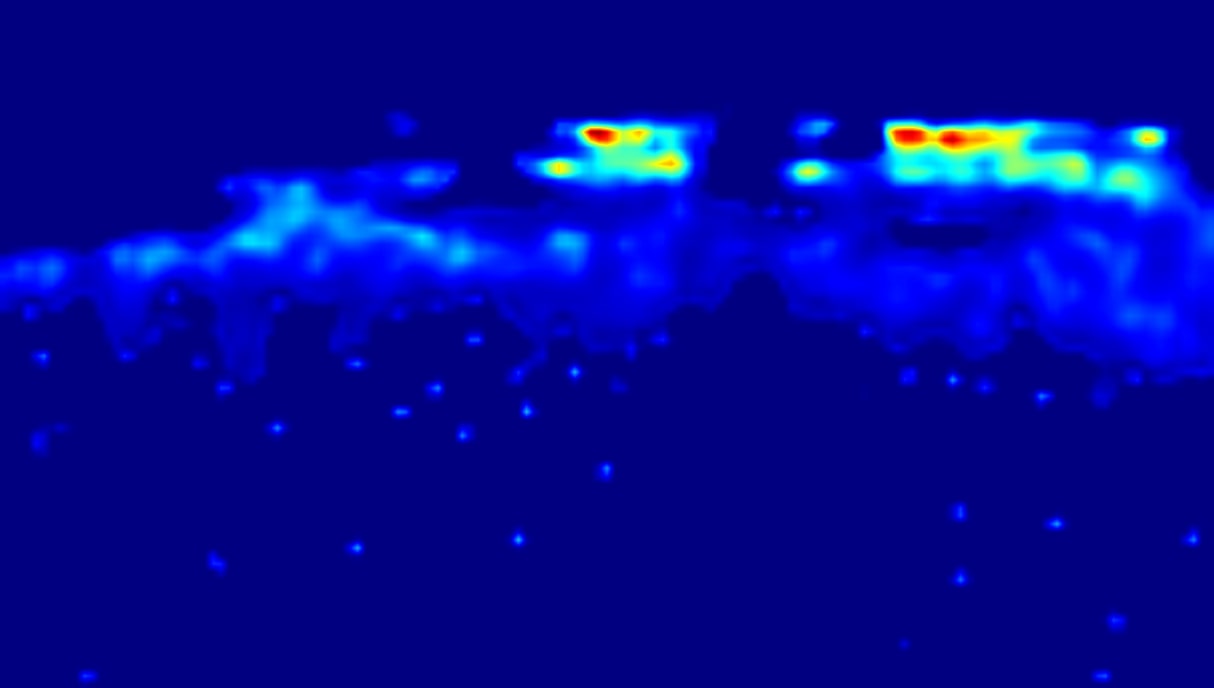} \hfill
        \includegraphics[width=0.195\textwidth]{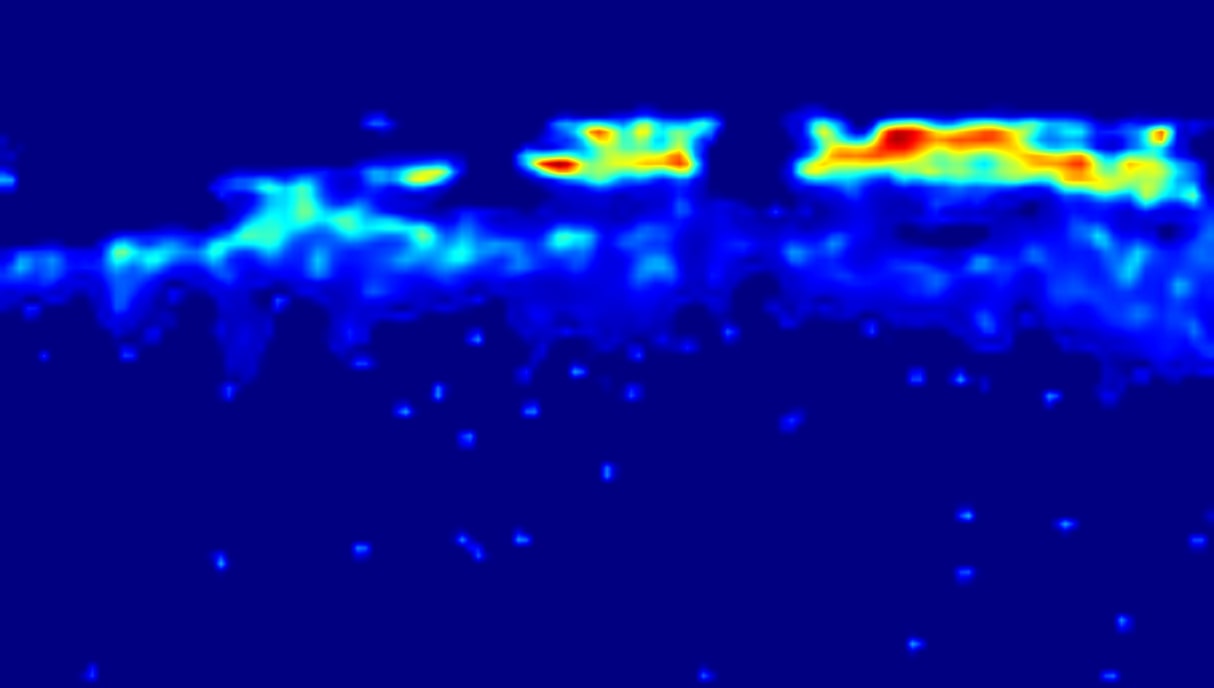}     
    \end{minipage}
    \\
    \begin{minipage}[c]{0.02\textwidth}
    \rotatebox[origin = c]{90}{}
    \end{minipage}    
    \begin{minipage}[c]{0.975\textwidth}
    \centering
        \includegraphics[width=0.195\textwidth]{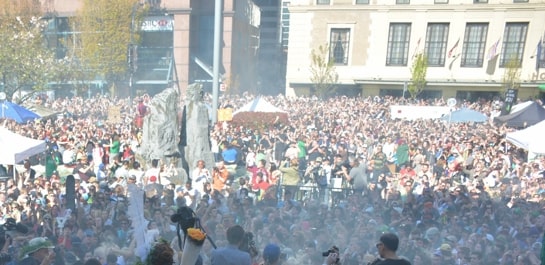} \hfill
        \includegraphics[width=0.195\textwidth]{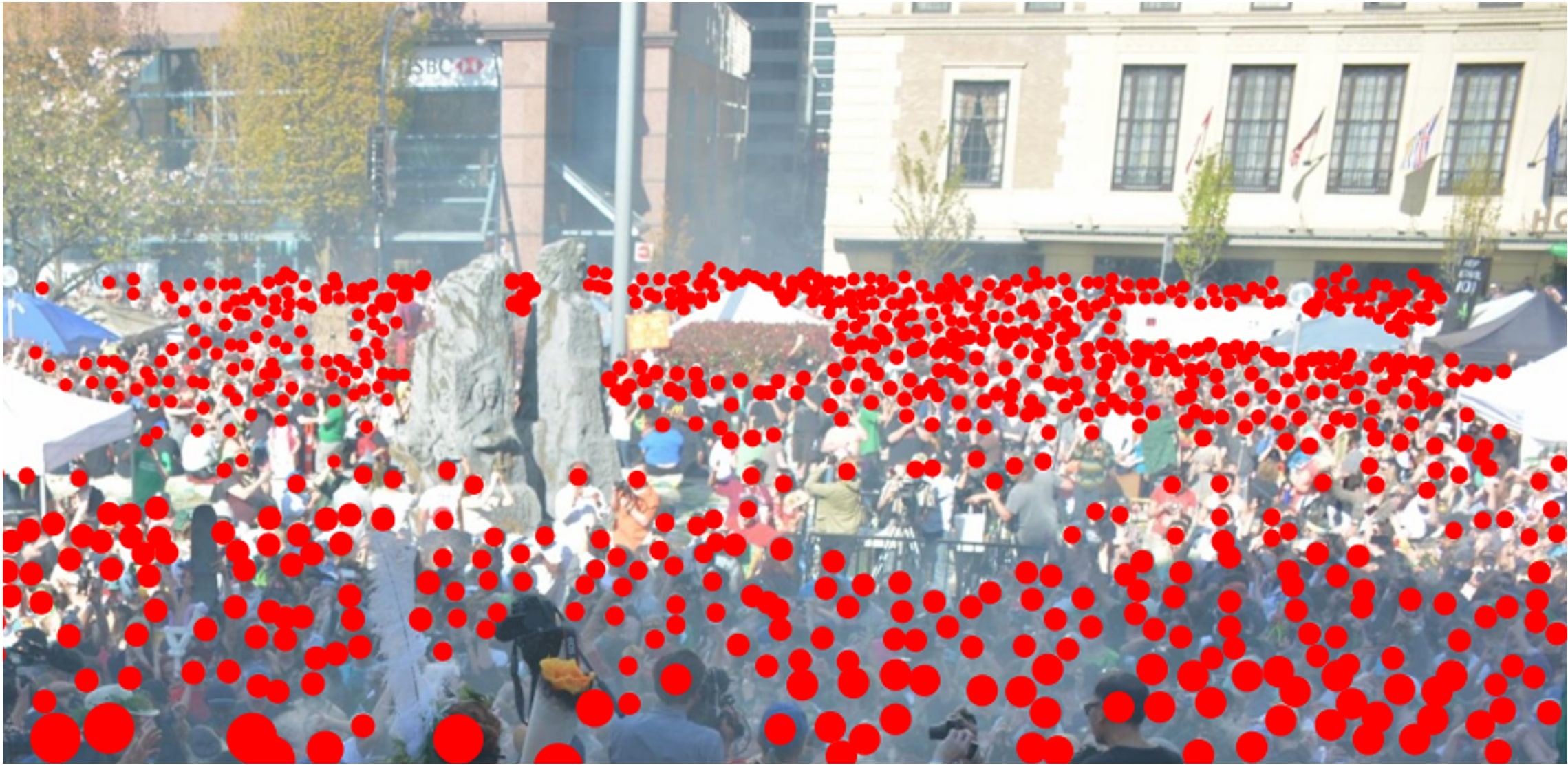} \hfill
        \includegraphics[width=0.195\textwidth]{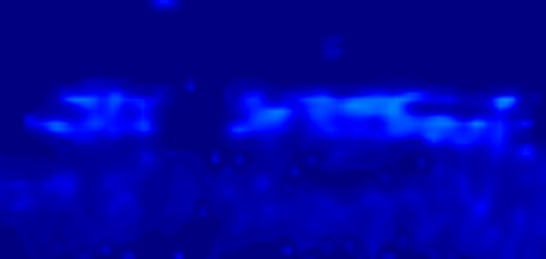} \hfill
        \includegraphics[width=0.195\textwidth]{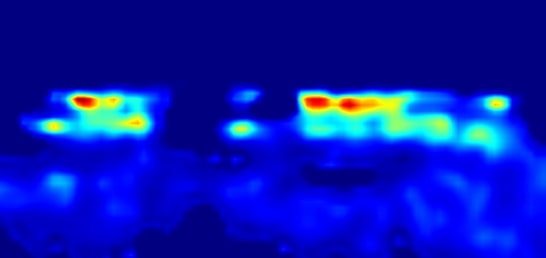} \hfill
        \includegraphics[width=0.195\textwidth]{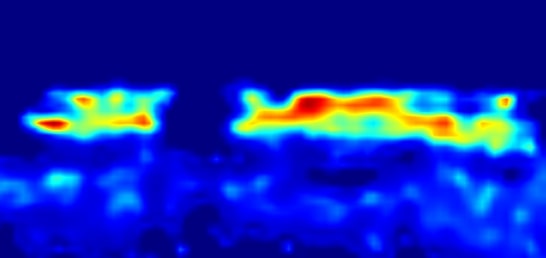}     
    \end{minipage}
    \\
    \hspace{0.006\textwidth}
     \begin{minipage}[c]{0.195\textwidth}
      \centering    
        \text{}
    \end{minipage}
    \begin{minipage}[c]{0.195\textwidth}
     \centering
        \text{Count=1134}
    \end{minipage}
         \begin{minipage}[c]{0.195\textwidth}
     \centering
        \text{Count=1406}
    \end{minipage}
         \begin{minipage}[c]{0.195\textwidth}
     \centering
        \text{Count=955}
    \end{minipage}
         \begin{minipage}[c]{0.195\textwidth}
     \centering
        \text{Count=1094}
    \end{minipage}
    \\
    \begin{minipage}[c]{0.02\textwidth}
    \rotatebox[origin = c]{90}{Snow}
    \end{minipage}
    \begin{minipage}[c]{0.975\textwidth}
    \centering
        \includegraphics[width=0.195\textwidth]{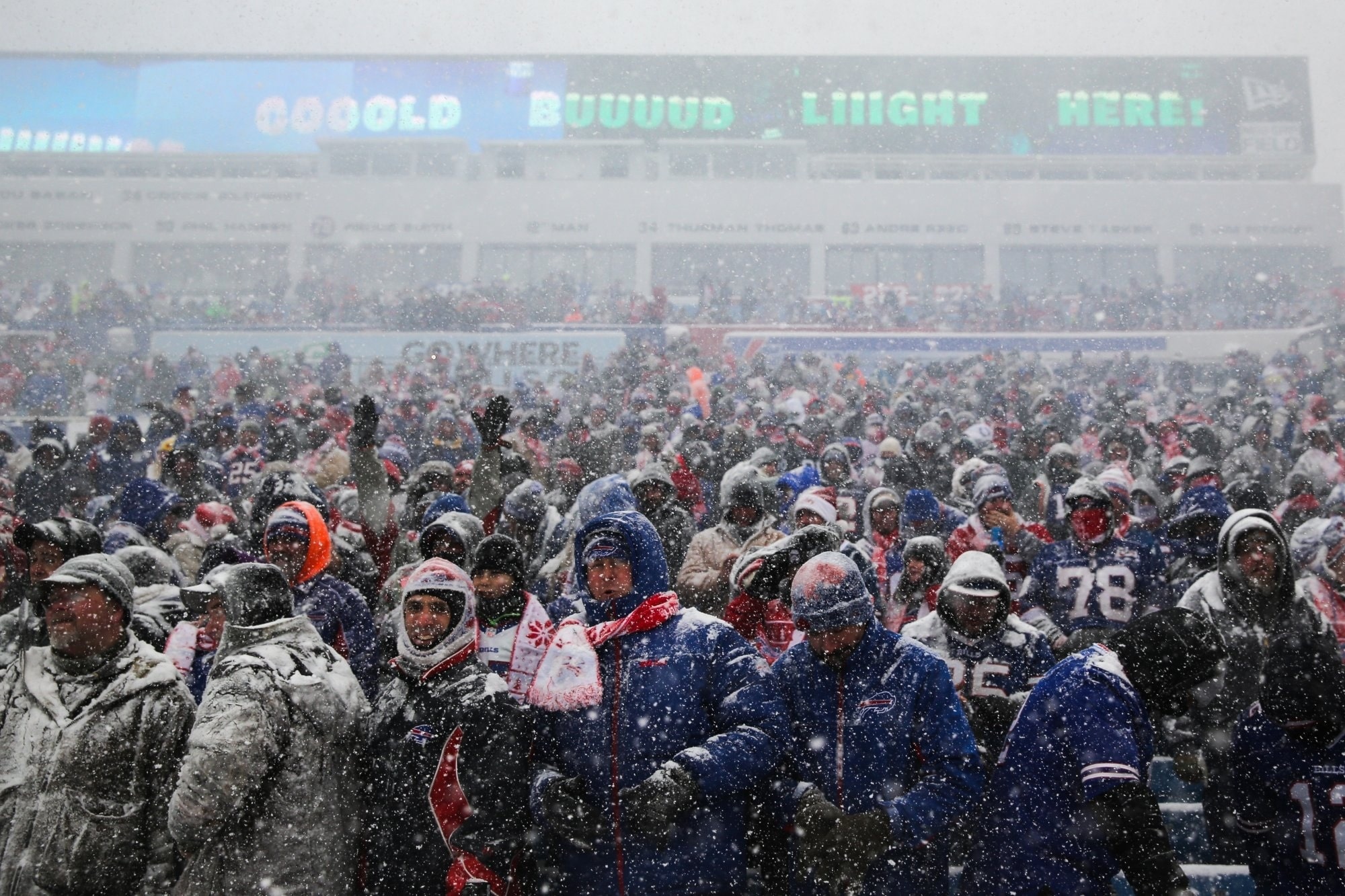} \hfill
        \includegraphics[width=0.195\textwidth]{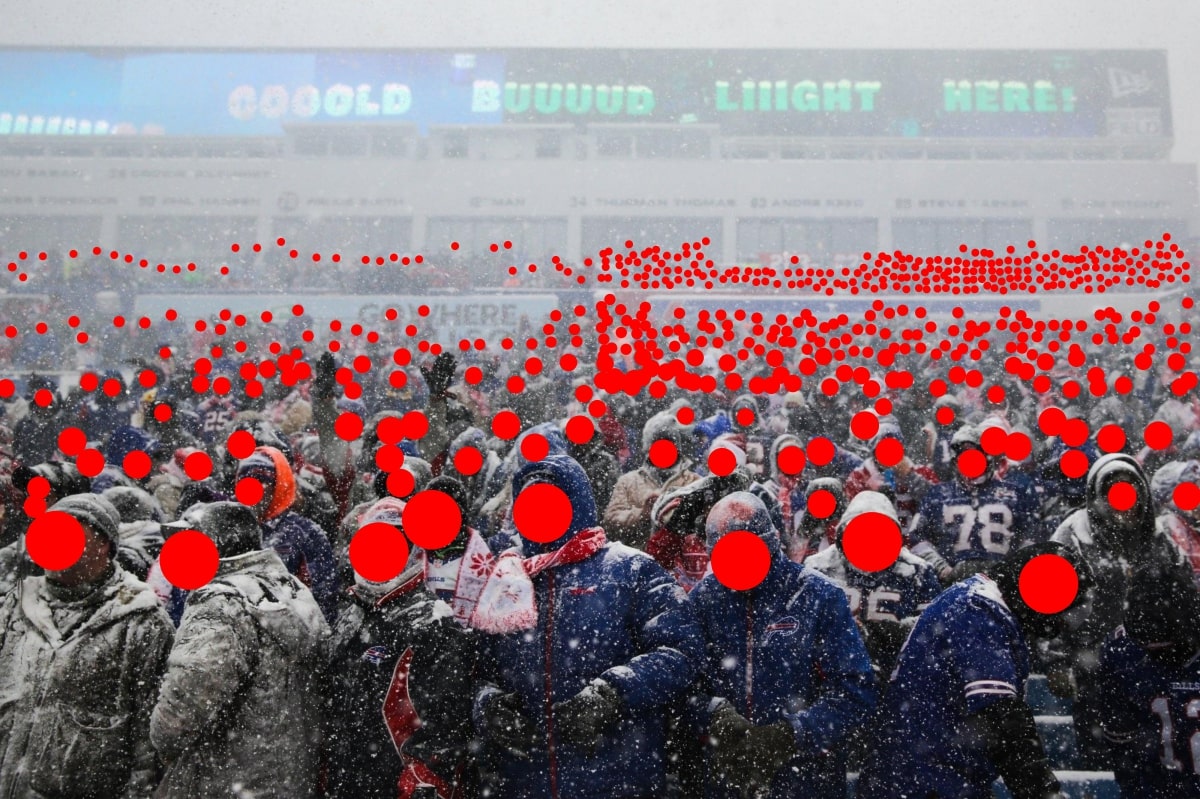} \hfill
        \includegraphics[width=0.195\textwidth]{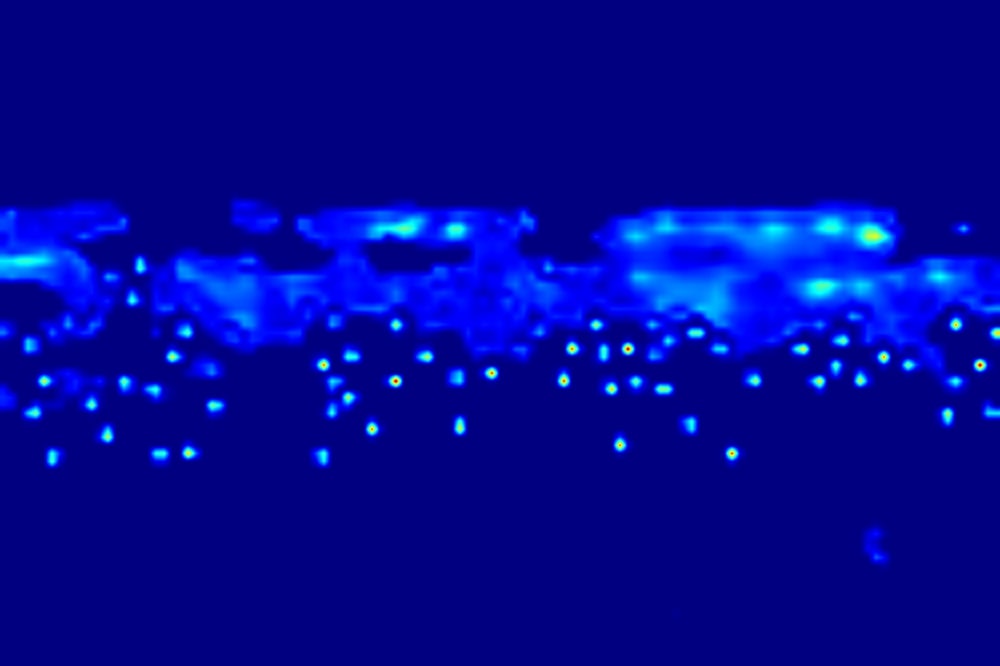} \hfill
        \includegraphics[width=0.195\textwidth]{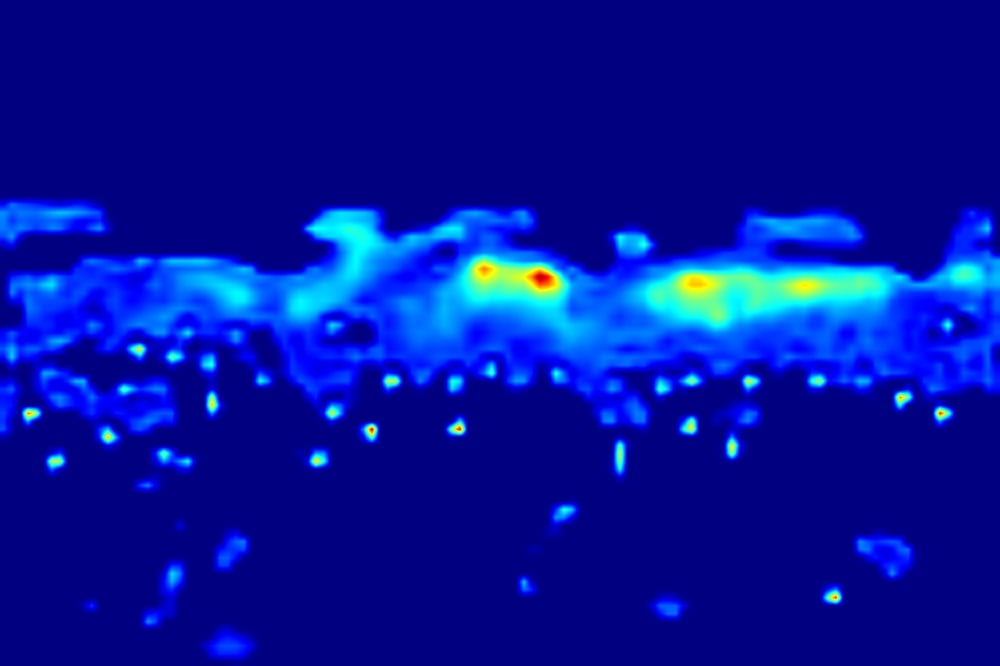} \hfill
        \includegraphics[width=0.195\textwidth]{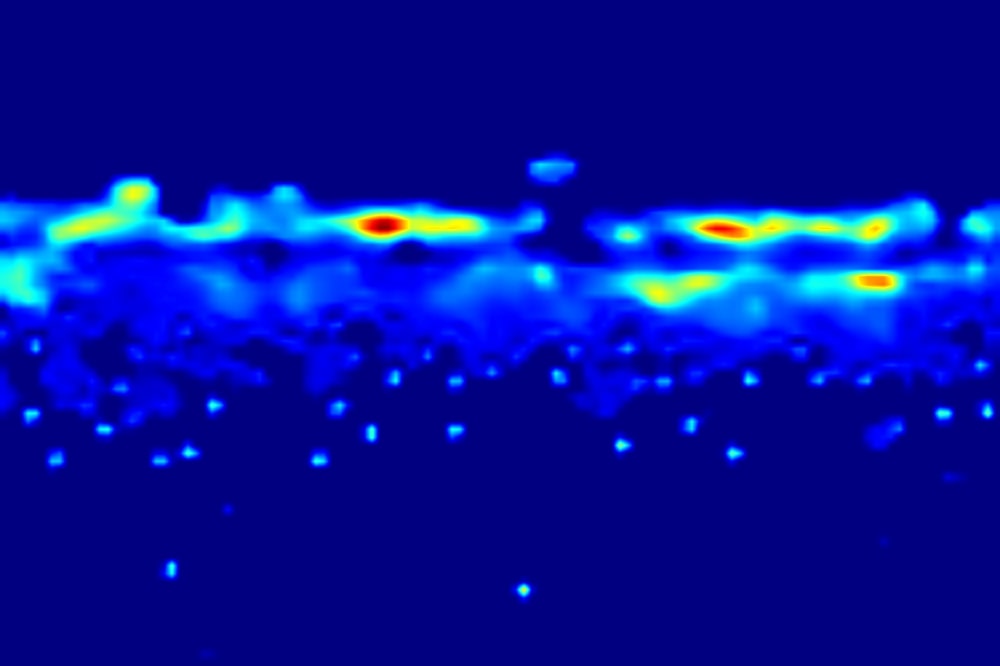}
    \end{minipage}
    \\
    \begin{minipage}[c]{0.02\textwidth}
    \rotatebox[origin = c]{90}{}
    \end{minipage}    
    \begin{minipage}[c]{0.975\textwidth}
    \centering
        \includegraphics[width=0.195\textwidth]{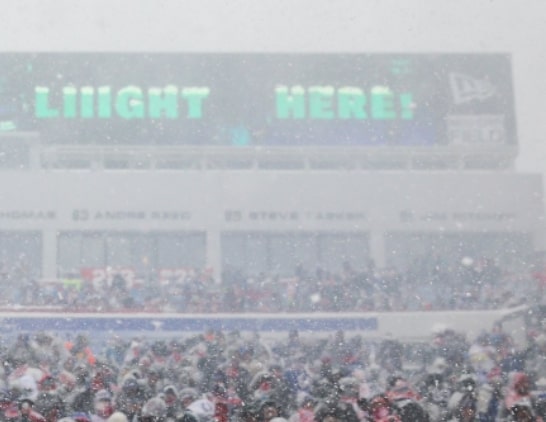} \hfill
        \includegraphics[width=0.195\textwidth]{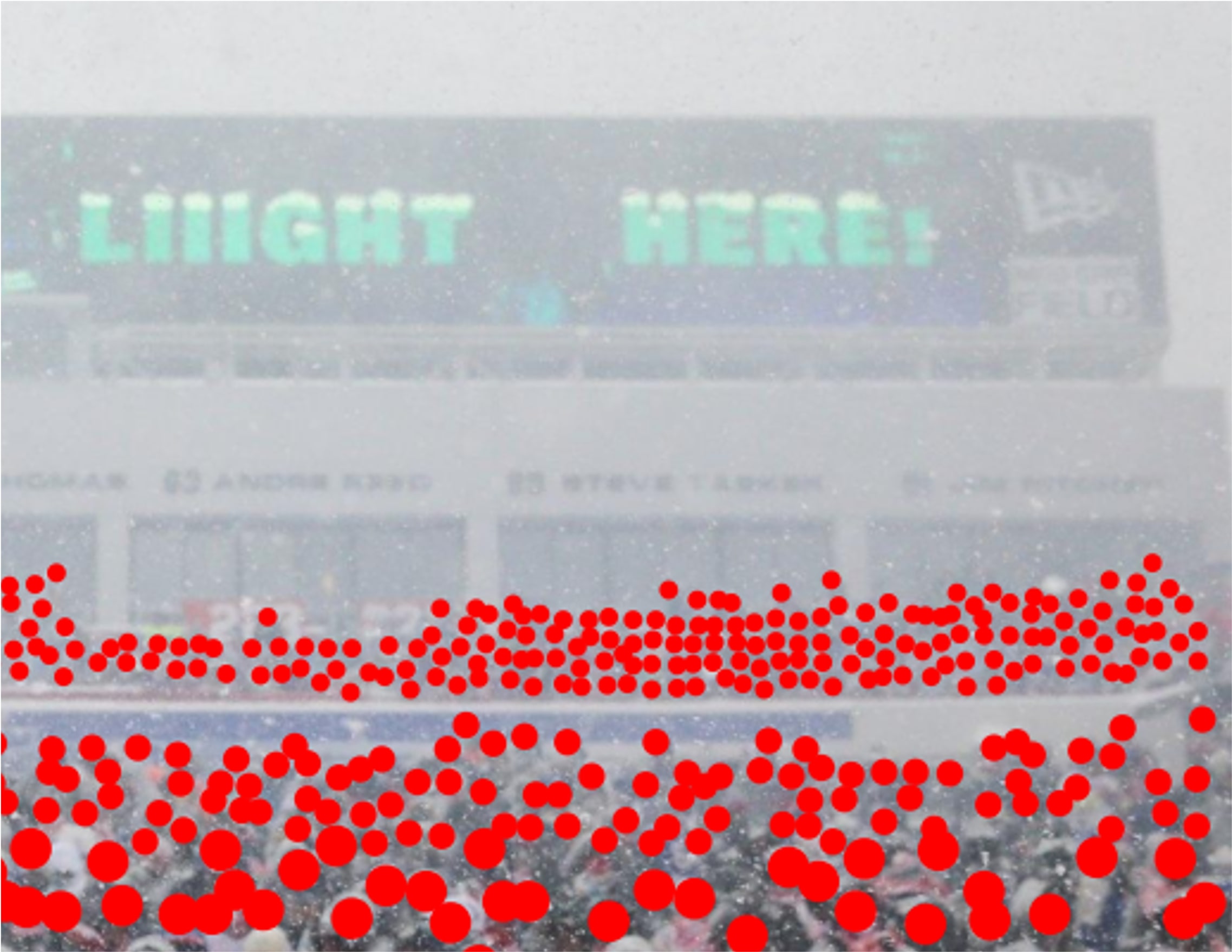} \hfill
        \includegraphics[width=0.195\textwidth]{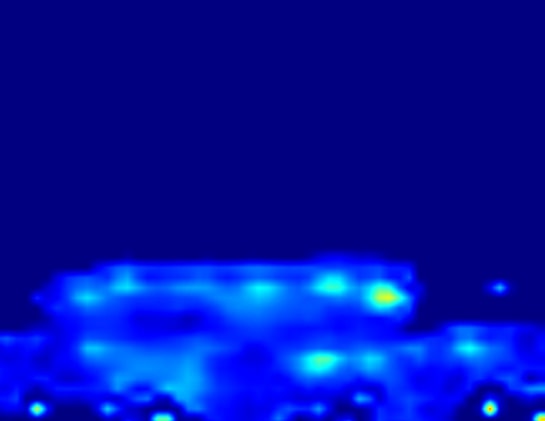} \hfill
        \includegraphics[width=0.195\textwidth]{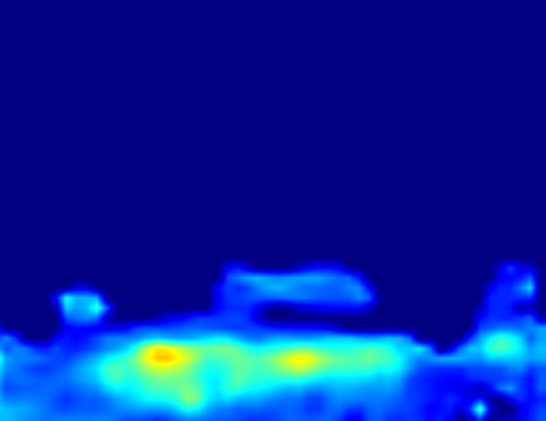} \hfill
        \includegraphics[width=0.195\textwidth]{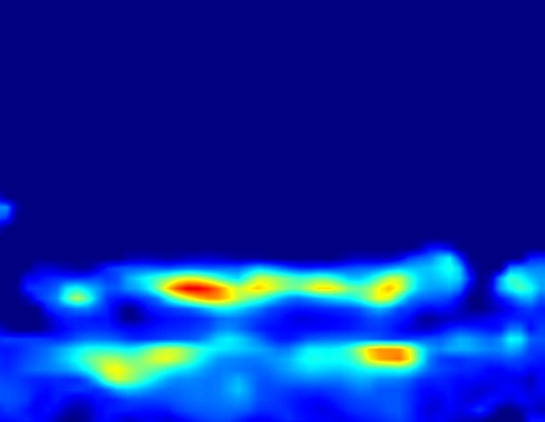}     
    \end{minipage}
    \\
    \hspace{0.006\textwidth}
     \begin{minipage}[c]{0.195\textwidth}
      \centering    
        \text{}
    \end{minipage}
    \begin{minipage}[c]{0.195\textwidth}
     \centering
        \text{Count=536}
    \end{minipage}
         \begin{minipage}[c]{0.195\textwidth}
     \centering
        \text{Count=279}
    \end{minipage}
         \begin{minipage}[c]{0.195\textwidth}
     \centering
        \text{Count=296}
    \end{minipage}
         \begin{minipage}[c]{0.195\textwidth}
     \centering
        \text{Count=482}
    \end{minipage}
    \\
    \begin{minipage}[c]{0.02\textwidth}
    \rotatebox[origin = c]{90}{Rain}
    \end{minipage}
    \begin{minipage}[c]{0.975\textwidth}
    \centering
        \includegraphics[width=0.195\textwidth]{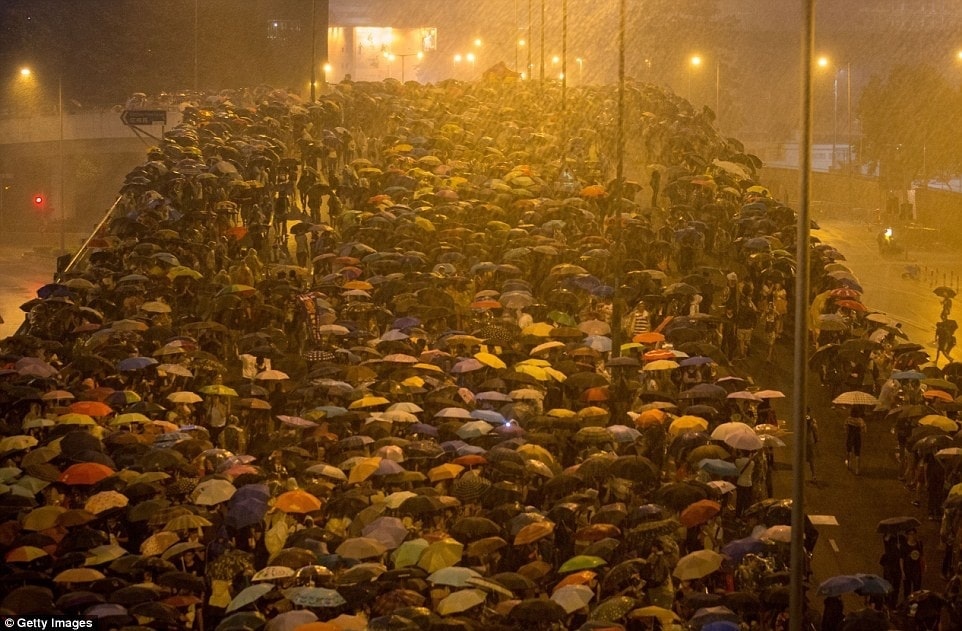} \hfill
        \includegraphics[width=0.195\textwidth]{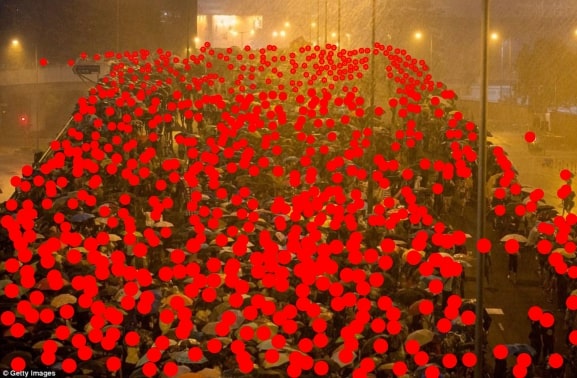} \hfill
        \includegraphics[width=0.195\textwidth]{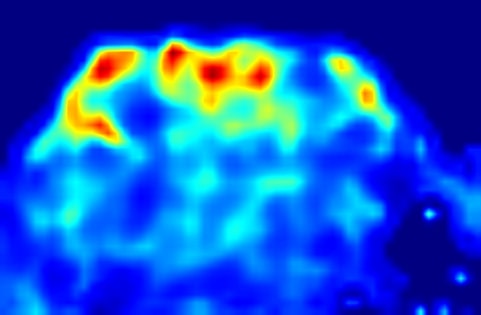} \hfill
        \includegraphics[width=0.195\textwidth]{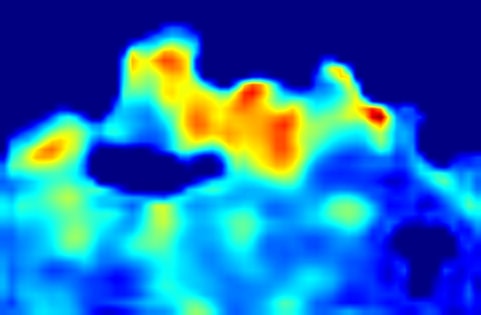} \hfill
        \includegraphics[width=0.195\textwidth]{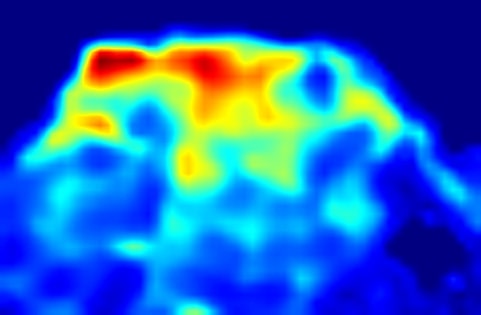}     
    \end{minipage}
    \\
    \begin{minipage}[c]{0.02\textwidth}
    \rotatebox[origin = c]{90}{}
    \end{minipage}    
    \begin{minipage}[c]{0.975\textwidth}
    \centering
        \includegraphics[width=0.195\textwidth]{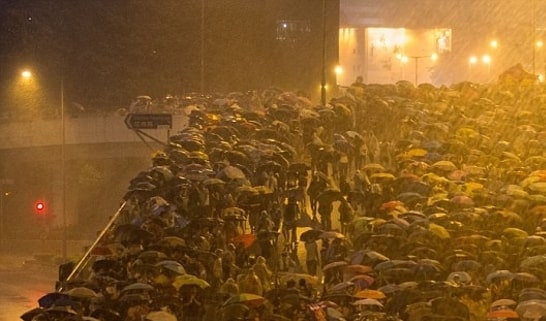} \hfill
        \includegraphics[width=0.195\textwidth]{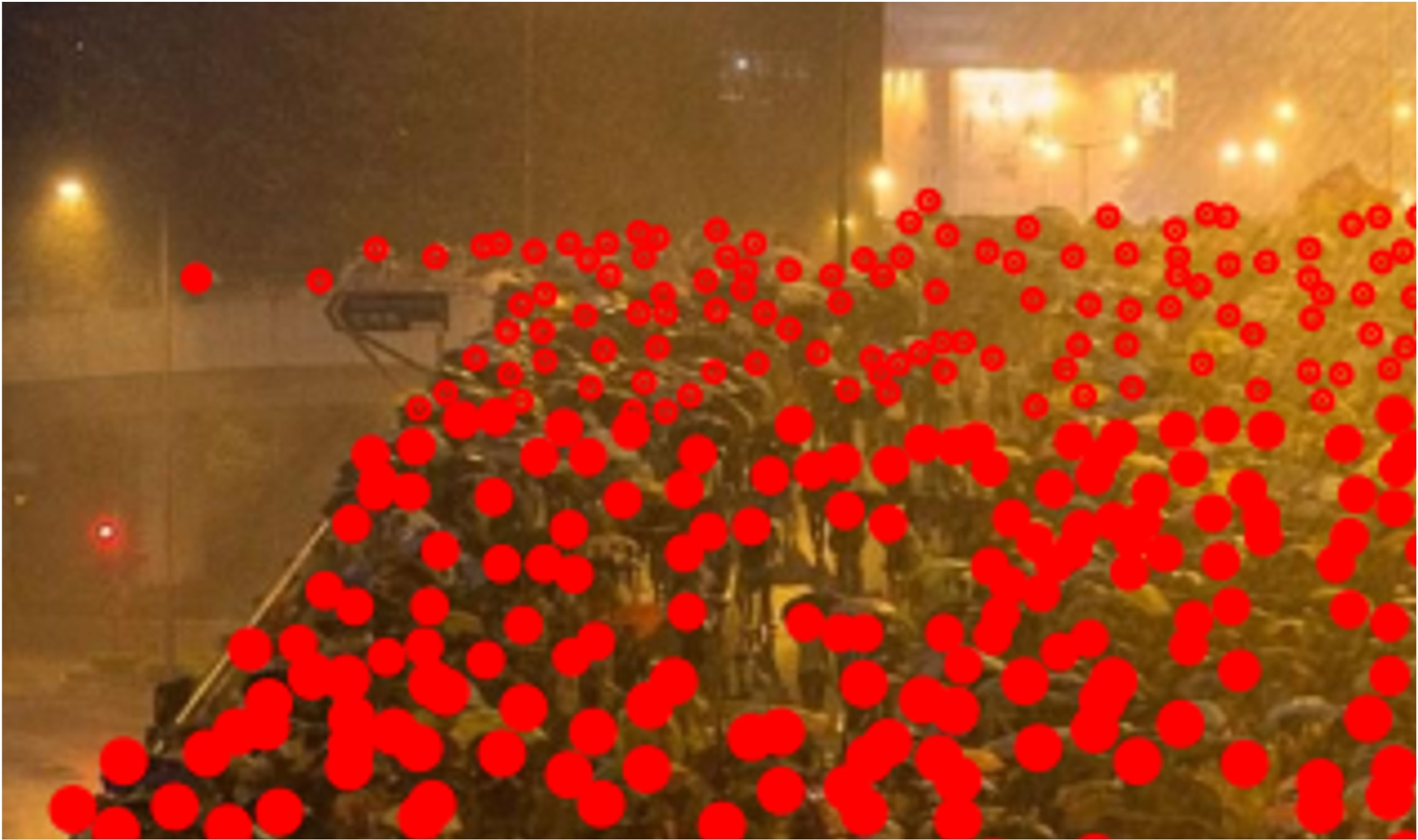} \hfill
        \includegraphics[width=0.195\textwidth]{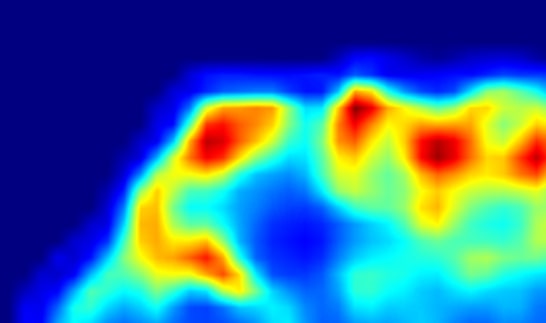} \hfill
        \includegraphics[width=0.195\textwidth]{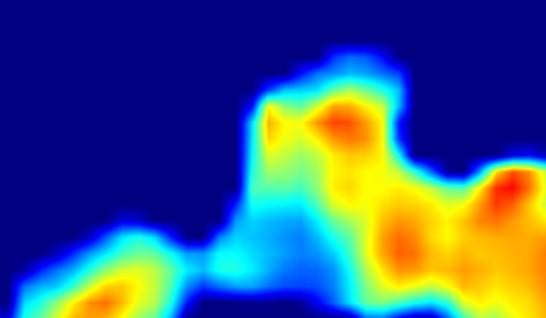} \hfill
        \includegraphics[width=0.195\textwidth]{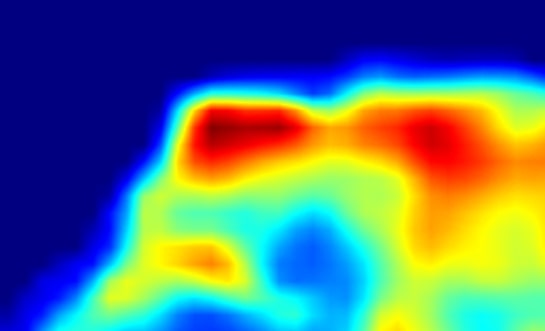}     
    \end{minipage}
    \\
    \hspace{0.006\textwidth}
     \begin{minipage}[c]{0.195\textwidth}
      \centering    
        \text{}
    \end{minipage}
    \begin{minipage}[c]{0.195\textwidth}
     \centering
        \text{Count=727}
    \end{minipage}
         \begin{minipage}[c]{0.195\textwidth}
     \centering
        \text{Count=583}
    \end{minipage}
         \begin{minipage}[c]{0.195\textwidth}
     \centering
        \text{Count=368}
    \end{minipage}
         \begin{minipage}[c]{0.195\textwidth}
     \centering
        \text{Count=684}
    \end{minipage}
    \\    
    \figmargin
\end{center}
    \caption{
        \textbf{Comparison of density maps of the proposed method and other baselines in adverse weather (i.e., haze, snow, rain).} The proposed method can predict more accurate density maps compared to the results estimated by other strategies.
    } 
    \label{fig:density}
\end{figure*}

\begin{figure}[t!]
\centering \includegraphics[width=0.47\textwidth,page=1]{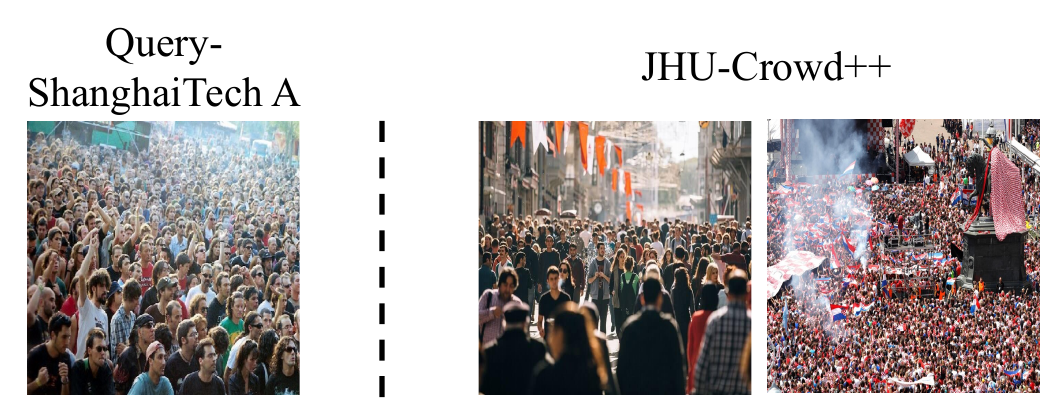}{}
\makeatother 
\caption{\textbf{Probe the weather type of an image of the ShanhaiTech A dataset from the JHU-Crowd++ dataset based on the weather-adaptive queries.} The learned weather prototypes and the weight vector can well represent the weather type of the input image from the unseen dataset.}
\label{fig:dg_visual}
\end{figure}

\begin{figure*}[t!]
\begin{center}
	\footnotesize
    \begin{minipage}[c]{0.95\textwidth}
    \centering
        \includegraphics[width=0.475\textwidth,page=1]{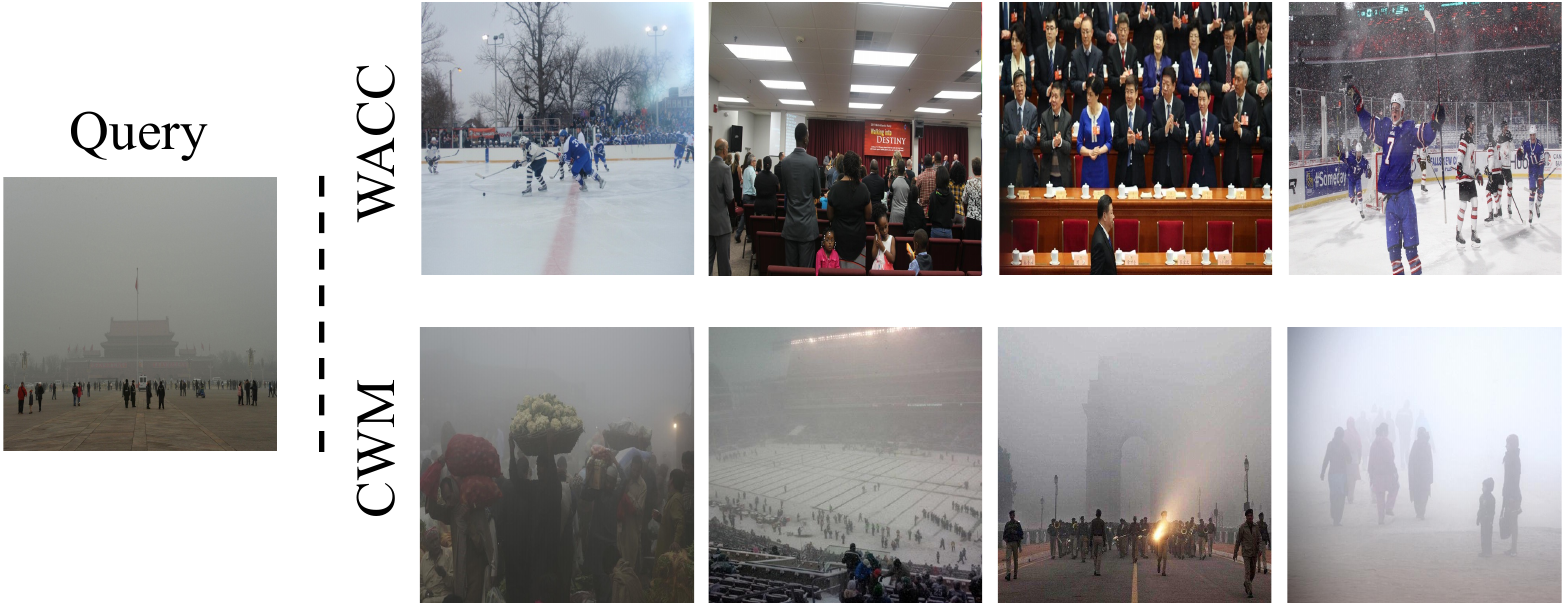}{}
        \hfill
        \includegraphics[width=0.475\textwidth,page=1]{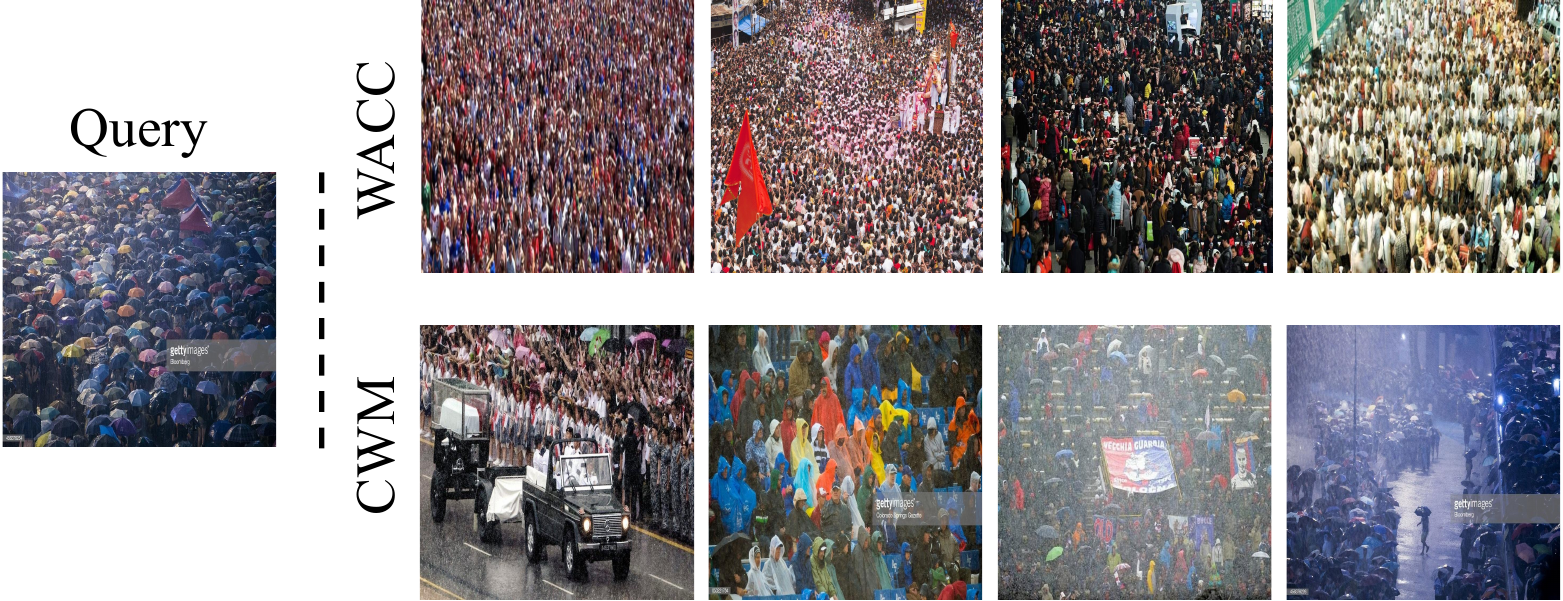}{}
    \end{minipage}
    \\
    \vspace{0.2cm}
    \begin{minipage}[c]{0.95\textwidth}
    \centering
        \includegraphics[width=0.475\textwidth,page=1]{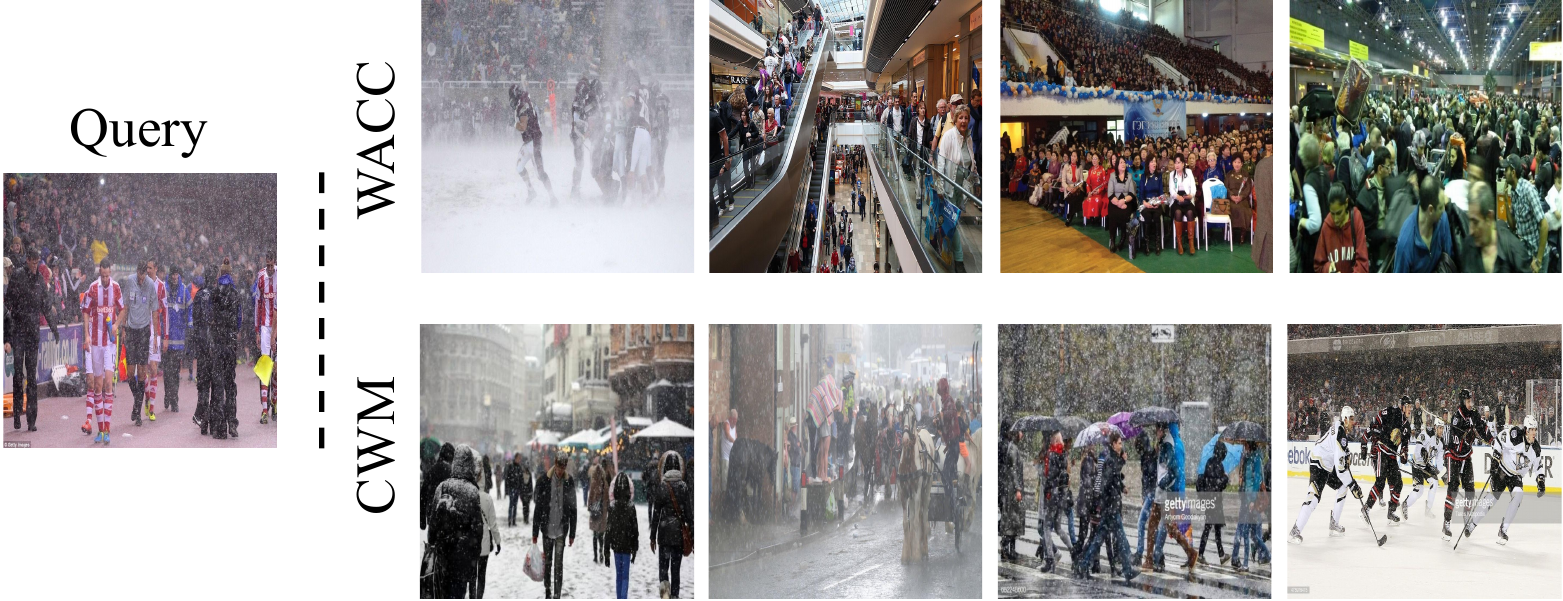}{}
 \hfill
        \includegraphics[width=0.475\textwidth,page=1]{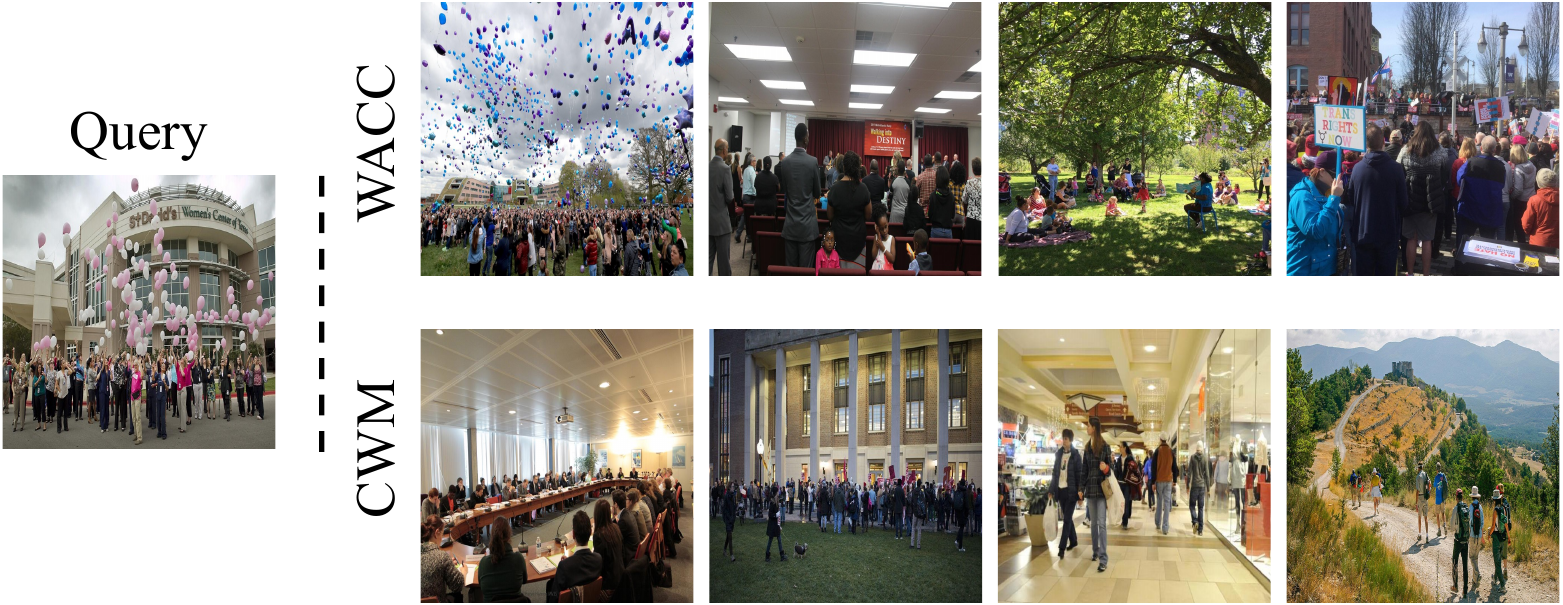}{}
    \end{minipage}
    \\    
    \figmargin
\end{center}
    \caption{\textbf{Visual comparison of the proposed contrastive weather-adaptive queries.} The proposed contrastive weather-adaptive queries can achieve a better representation of the weather type. We term the proposed contrastive weather-adaptive module as 'CWM'.} 
 \label{fig:anchor_query}
\end{figure*}

\subsection{Performance Evaluation}
We evaluate our method against the state-of-the-art approaches including SFCN~\cite{Wang2019LearningFS}, BL~\cite{ma2019bayesian}, LSCCNN~\cite{sam2020locate}, CG-DRCN~\cite{sindagi2020jhu}, UOT~\cite{ma2021learning}, S3~\cite{lin2021direct}, GL~\cite{wan2021generalized}, ChfL~\cite{shu2022crowd}, CLTR~\cite{liang2022end}, MAN~\cite{lin2022boosting}, and GauNet~\cite{cheng2022rethinking}.
 We present the evaluation results of our AWCC-Net on the JHU-Corwd++ dataset for adverse weather crowd counting since only this dataset contains the annotations of bad weather. 
 We also report the performance on the ShanghaiTech A, UCF-QNRF, and NWPU-CROWD datasets.
\noindent \smallskip\\
\textbf{Evaluation on the JHU-Crowd++ dataset.} We use two settings in the experiments:
\begin{compactenum}
\item \textbf{Vanilla crowd counting}: We apply vanilla crowd counting methods trained with the training set of the JHU-Crowd++ dataset.
\item \textbf{Two-stage strategy}: We use the Unified model~\cite{chen2022learning} to restore image content under adverse weather and then apply vanilla crowd counting methods to estimate the counts. We denote this strategy with '-U'. 
Note the crowd counting models are not fine-tuned by restored images. 
For fair comparisons, we adopt these restored images to finetune the crowd counting models. We denote this strategy with '-UF'. 
\end{compactenum}

\tabref{tab:performance} shows that, although existing methods achieve state-of-the-art results in clear scenes, they do not perform well in adverse weather scenes. 
Second, the two-stage strategy may not help the improvement of the performance compared to the vanilla crowd counting approaches since the goal of the image restoration process is not the same as counting. 
This is because the image restoration methods are designed for restoring content rather than crowd counting, and thus the restored results may not suitable for crowd computation. 
Third, our method performs favorably against state-of-the-art approaches in adverse weather. 
For the clear scenes, our method can achieve comparable performance in MAE. 
In terms of average performance, our method can achieve the first place in both MAE and MSE. 
Overall, the proposed method is effective and robust when the input images are degraded by bad weather while it can retain the performance of clear scenes. 

We also demonstrate the density maps under different adverse weather predicted by the proposed method and other algorithms in \figref{fig:density}.
The results show that the proposed method can compute the more accurate density distribution and counts of crowds under bad weather. 
\noindent \smallskip\\
\textbf{Evaluation on other datasets.} We present the evaluation results of our algorithm on other datasets in \tabref{tab:performance_other}. 
As these images do not contain annotations of weather types in the images, we can only compare them in the averaged performance. 
The results indicate that the proposed method performs favorably in the JHU-Crowd++ and UCF-QNRF datasets while comparable performance in the ShanghaiTech A and NWPU-CROWD datasets.
\noindent \smallskip\\
\textbf{Generalization to other datasets.} We analyze the generalization ability of our method on other datasets. 
Specifically, we train the AWCC-Net and other baselines on the JHU-Crowd++ dataset and directly test on the ShanghaiTech A, and UCF-QNRF datasets. 
\tabref{tab:generalization} shows that the proposed method achieves the best performance in terms of MAE and MSE. 
Moreover, we use an image in the ShanghaiTech A dataset and search for the most similar two images in the JHU-Crowd++ dataset. 
The two images whose weather queries from the training dataset are most similar to the weather queries of the input from the unseen dataset are presented in \figref{fig:dg_visual}. 
Our model can locate the images with the similar weather type in the training set based on the combination of the weight vector and the weather bank. Thus, the proposed weather-aware crowd counting mechanism and the weather-adaptive queries can benefit our network to be robust to the weather types in the unseen datasets.

\begin{table}[t!]
\centering

\begin{tabular}{ccccc} 
\toprule
\textbf{Dataset}  & \multicolumn{2}{c}{\textbf{ShanghaiTechA}} & \multicolumn{2}{c}{\textbf{UCF-QNRF}}  \\ 
\hline
\textbf{Method}   & MAE & MSE            & MAE & MSE       \\
\hline\hline
BL~\cite{ma2019bayesian}    &  102.6   &  186.1     &  164.7   &  297.0         \\
GL~\cite{wan2021generalized}&  103.7   &  188.8     &  155.0   &  283.9         \\
MAN~\cite{lin2022boosting}  &  81.2    &  160.4     &  132.8   &  241.9         \\
\textbf{AWCC-Net}           &  \textbf{73.2}    &  \textbf{132.4}     &  \textbf{120.9}   &  \textbf{216.7}         \\
\bottomrule
\end{tabular}
\\
\caption{\textbf{Analysis on the generalization ability of the proposed AWCC-Net.} Our method presents better generalization ability on other datasets.}
\label{tab:generalization}
\end{table}

\begin{table}[t!]
\centering
\scalebox{1.0}{
\begin{tabular}{cccc|cc} 
\toprule
\multicolumn{4}{c|}{\textbf{Module}} & \multicolumn{2}{c}{\textbf{Metric}}  \\ 
\hline
WACC & IWQ & $\mathcal{L}_{Con}$ & $\mathcal{L}_{CP}$ & MAE & MSE \\
\hline\hline
- & - & - & -  &  60.7 & 230.4   \\
\checkmark & - & - & - & 60.1 & 229.5 \\
\checkmark & \checkmark & - & - & 59.3 & 226.1 \\
\checkmark & \checkmark & \checkmark & -   &  53.1   &  211.4 \\
\checkmark & \checkmark & \checkmark & \checkmark & \textbf{52.3} & \textbf{207.2}  \\
\bottomrule
\end{tabular}}
\\
\caption{\textbf{Analysis on alternatives of AWCC-Net.} Note that we term the input-dependent weather queries as 'IWQ'. We demonstrate that the contrastive weather-adaptive queries can improve the performance of crowd counting effectively.}
\label{tab:ablation}
\end{table}

\begin{table}[t!]
\centering
\scalebox{1.0}{
\begin{tabular}{ccc} 
\toprule
\textbf{Strategy }      & MAE & MSE  \\ 
\hline\hline
AWCC-Net-Label &      56.1        &     220.3          \\
AWCC-Net       &      \textbf{52.3}        &     \textbf{207.2}          \\
\bottomrule
\end{tabular}}
\\
\caption{\textbf{Analysis on the necessity of using weather label in the AWCC-Net.} We demonstrate adopting weather labels may potentially degrade the performance of crowd counting.}
\label{tab:necessity}
\end{table}

\subsection{Ablation Study}
We evaluate the effectiveness of weather-aware crowd counting and contrastive weather-adaptive queries using the JHU-Crowd++ dataset.
\noindent \smallskip\\
\textbf{Effectiveness of Weather-aware Crowd Counting.} 
We evaluate the proposed WACC mechanism in \tabref{tab:ablation}. 
With only the proposed WACC architecture (MAE of 60.1 and MSE of 229.5), moderate improvements over the baseline model (MAE of 60.7 and MSE of 230.4) are achieved.  
As stated in \ref{sec:WACC}, it is due to the learning of weather-unrelated information in the weather queries.
\noindent \smallskip\\
\textbf{Effectiveness of Contrastive Weather-adaptive Module.} 
The contrastive weather-adaptive module facilitates the weather queries in WACC to be effective for crowd counting, as shown in~\tabref{tab:ablation}. 
The proposed input-dependent weather queries and contrastive loss can improve the performance of WACC effectively since the learned weather queries are constrained to learn weather information effectively.
Moreover, the compact prototype loss facilitates learning compact prototypes in the weather bank. 

We validate the effectiveness of the proposed module visually in \figref{fig:anchor_query}. 
We compute the weather queries on images outside of the training set.
Then, we randomly pick images of different weather types (i.e., snow, rain, and haze) and clear scenes as queries. 
Based on the chosen weather queries, we identify the images with the top-4 smallest distance from the remaining weather queries.
The results show that the proposed mechanism can represent the various weather types effectively while the results without the proposed mechanism may find the images with different weather types since the learned weather queries can not well represent weather information.
\noindent \smallskip\\
\textbf{Necessity of Weather Annotations.} We analyze the performance of adopting annotations of weather type and the proposed learning weather queries strategy in the AWCC-Net. 
We construct a baseline based on the proposed AWCC-Net and leverage the annotations of weather types as guidance. 
Specifically, we remove the weight vector in the contrastive weather-adaptive module and set four weather prototypes in the weather bank since the JHU-Crowd++ dataset only contains annotations of haze, rain, snow, and clear scenes. 
We manually adopt one of these prototypes as the queries in the transformer according to the weather label of the input image. 
The results are reported in~\tabref{tab:necessity} where
the aforementioned baseline is denoted as 'AWCC-Net-Label'. 
The results show that using annotations of weather type in our method may degrade the performance of crowd counting since human-defined weather labels may potentially be mislabeled and contain the weather ambiguity problem. 
In contrast, our network does not adopt the human label of the weather types and learns the weather prototypes automatically, which shows the better performance in adverse weather crowd counting problems.

\section{Conclusion}
In this paper, we propose the AWCC-Net model to address the crowd counting problem under adverse weather. 
We introduce the weather query mechanism to the crowd counting network, which enables our network to learn weather-aware feature extraction.  
To enforce the weather queries to learn the weather information effectively, we propose a module and contrastive loss to learn weather-adaptive queries for robust crowd counting. 
Moreover, the compact prototype loss is proposed to improve the model performance. 
Extensive experimental results show that the proposed method performs favorably against the state-of-the-art methods in both adverse weather and clear images.

\section{Acknowledgement}
We thank to National Center for High-performance Computing (NCHC) for providing computational and storage resources.

\clearpage

\appendix

\renewcommand{\theequation}{S.\arabic{equation}}
\renewcommand\thefigure{S.\arabic{figure}}
\renewcommand\thetable{S.\arabic{table}}
\setcounter{equation}{0}
\setcounter{figure}{0}
\setcounter{table}{0}

{
\onecolumn
\begin{center}
\textbf{\large Supplemental Materials}
\end{center}
\section{More Experimental Results}
We evaluate our method with more methods including MCNN~\cite{Zhang2016SingleImageCC}, CSR-Net~\cite{Li2018CSRNetDC}, SA-Net~\cite{Cao2018ScaleAN}, and NoisyCC~\cite{Wan2020ModelingNA} on the ShanghaiTech~\cite{zhang2016single}, UCF-QNRF~\cite{idrees2018composition}, JHU-Crowd++~\cite{sindagi2020jhu}, and NWPU-CROWD~\cite{wang2020nwpu} datasets. The results are demonstrated in~\tabref{tab:supp_performance_other}. The proposed AWCC-Net can achieve the best performance in UCF-QNRF and JHU-Crowd++ while conducting comparable performance in the ShanghaiTechA and NWPU-CROWD datasets. Moreover, we present the density maps under different adverse weather and clear scene predicted by the AWCC-Net and other algorithms in~\figref{fig:supp_density}. The results show that the AWCC-Net can predict the more accurate density distribution and counts of crowds under bad weather and clear scenes.

\section{Implementation Details}
In Table 1 of the regular paper, we compare our methods with several baselines. The results of 'BL-U', 'BL-UF', 'GL', 'GL-U' and 'GL-UF' are retrained based on their original setting and official implementation since they do not provide pre-trained weights on the JHU-Crowd++ dataset. The results of other methods are directly reported from their original papers or the paper of the JHU-Crowd++ dataset~\cite{sindagi2020jhu}.

\begin{table*}[htb]
\centering
\small
\scalebox{0.95}{
\centering
\begin{tabular}{ccccccccc} 
\toprule
\multicolumn{1}{c}{\textbf{Dataset}} & \multicolumn{2}{c}{\textbf{ShanghaiTechA}} & \multicolumn{2}{c}{\textbf{UCF-QNRF}}  & \multicolumn{2}{c}{\textbf{JHU-Crowd++}}& \multicolumn{2}{c}{\textbf{NWPU-CROWD}}  \\ 
\hline
\multicolumn{1}{c}{\textbf{Method}}  & MAE & \multicolumn{1}{c}{MSE}     & MAE & \multicolumn{1}{c}{MSE} & MAE & MSE & MAE & MSE                 \\ 
\hline\hline

MCNN~\cite{Zhang2016SingleImageCC} & 110.2 & 173.2 & 277.0 & 426.0 & 188.9 & 483.4 & 232.5 & 714.6 \\
CSRNet~\cite{Li2018CSRNetDC} & 68.2 & 115.0 & - & - & 85.9 & 309.2 & 121.3 & 387.8 \\
SANet~\cite{Cao2018ScaleAN} & 67.0 & 104.5 & - & - & 91.1 & 320.4 & 190.6 & 491.4 \\

SFCN~\cite{Wang2019LearningFS} & 64.8 & 107.5 & 102.0 & 171.4 & 77.5 & 297.6 & 105.7 & 424.1 \\
BL~\cite{ma2019bayesian} & 62.8 & 101.8 & 88.7 & 154.8 & 75.0 & 299.9 & 105.4 & 454.2 \\
LSCCNN~\cite{sam2020locate} & 66.5 & 101.8 & 120.5 & 218.2 & 112.7 & 454.4 & - & - \\ 
CG-DRCN-VGG16~\cite{sindagi2020jhu} & 64.0 & 98.4 & 112.2 & 176.3 & 82.3 & 328.0& - & - \\
CG-DRCN-Res101~\cite{sindagi2020jhu} & 60.2 & 94.0 & 95.5 & 164.3 & 71.0 & 278.6 & - & - \\

DM-Count~\cite{wang2020distribution} & 59.7 & 95.7 & 85.6 & 148.3 & - & - & 88.4 & 388.6 \\ 
NoisyCC~\cite{Wan2020ModelingNA} & 61.9 & 99.6 & 85.8 & 150.6 & - & - & 96.9 & 534.2 \\

UOT~\cite{ma2021learning} & 58.1 & 95.9 & 83.3 & 142.3 & 60.5 & 252.7 & 87.8 & 387.5 \\
S3~\cite{lin2021direct} & 57.0 & 96.0 & 80.6 & 139.8 & 59.4 & 244.0 & 83.5 & 346.9 \\
GL~\cite{wan2021generalized} & 61.3 & 95.4 & 84.3 & 147.5 & 59.9 & 259.5 & 79.3 & 346.1 \\
ChfL~\cite{shu2022crowd} & 57.5 & 94.3 & 80.3 & 137.6 & 57.0 & 235.7 & 76.8 & 343.0 \\
CLTR~\cite{liang2022end} & 56.9 & 95.2 & 85.8 & 141.3 & 59.5 & 240.6 & \textbf{74.3} & 333.8 \\
MAN~\cite{lin2022boosting}  & 56.8 & \underline{90.3} & \underline{77.3} & \underline{131.5} & \underline{53.4} & \underline{209.9} & 76.5 & \textbf{323.0} \\
GauNet~\cite{cheng2022rethinking} & \textbf{54.8} & \textbf{89.1} & 81.6 & 153.7 & 58.2 & 245.1 & - & - \\
\rowcolor{LightCyan}
\textbf{AWCC-Net} & \underline{56.2} & 91.3 & \textbf{76.4} & \textbf{130.5} & \textbf{52.3} & \textbf{207.2}& \underline{74.4} & \underline{329.1} \\
\bottomrule
\end{tabular}
 }
 \vspace{1mm}
\caption{\textbf{Quantitative comparison on the ShanghaiTech A~\cite{zhang2016single}, UCF-QNRF~\cite{idrees2018composition}, JHU-Crowd++~\cite{sindagi2020jhu}, and NWPU-CROWD~\cite{wang2020nwpu} datasets with existing methods.} The words with \textbf{boldface} indicate the best results, and those with \underline{underline} indicate the second-best results.}
\label{tab:supp_performance_other}
\end{table*}

\begin{figure*}[t!]
\begin{center}
	\footnotesize
    \hspace{0.006\textwidth}
     \begin{minipage}[c]{0.195\textwidth}
      \centering    
        \text{Input}
    \end{minipage}
    \begin{minipage}[c]{0.195\textwidth}
     \centering
        \text{Ground Truth}
    \end{minipage}
         \begin{minipage}[c]{0.195\textwidth}
     \centering
        \text{MAN~\cite{lin2022boosting}}   
    \end{minipage}
         \begin{minipage}[c]{0.195\textwidth}
     \centering
        \text{Unified~\cite{chen2022learning}+MAN~\cite{lin2022boosting}}    
    \end{minipage}
         \begin{minipage}[c]{0.195\textwidth}
     \centering
        \text{AWCC-Net}
    \end{minipage}
    \\
    \begin{minipage}[c]{0.02\textwidth}
    \rotatebox[origin = c]{90}{Haze}
    \end{minipage}
    \begin{minipage}[c]{0.975\textwidth}
    \centering
        \includegraphics[width=0.195\textwidth]{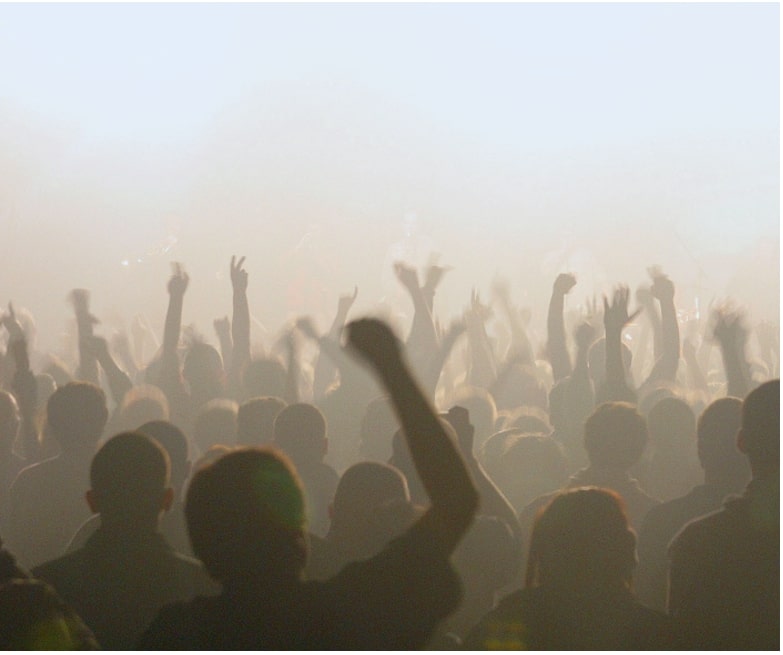} \hfill
        \includegraphics[width=0.195\textwidth]{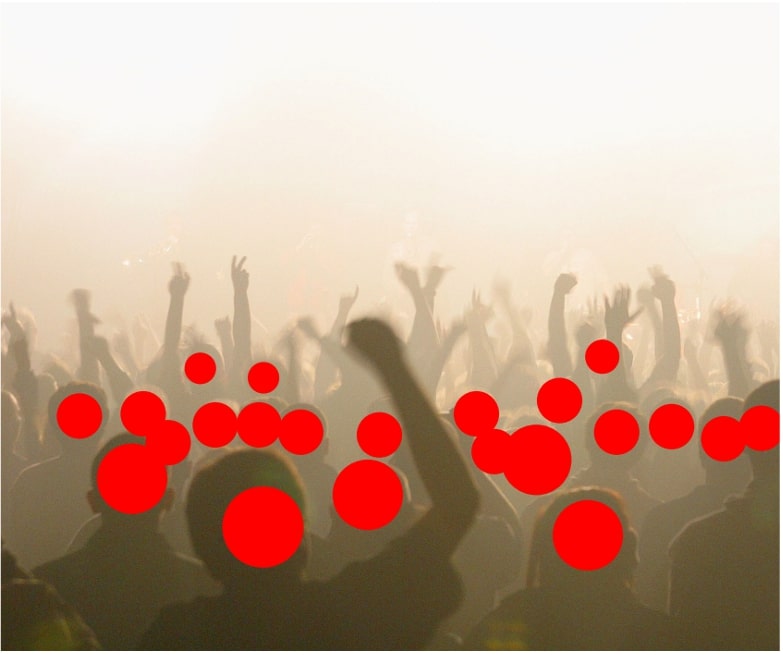} \hfill
        \includegraphics[width=0.195\textwidth]{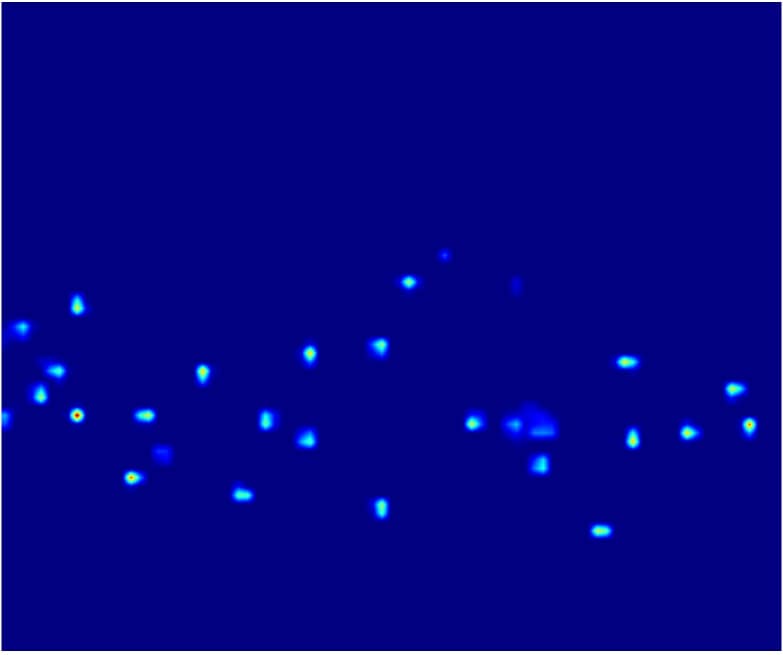} \hfill
        \includegraphics[width=0.195\textwidth]{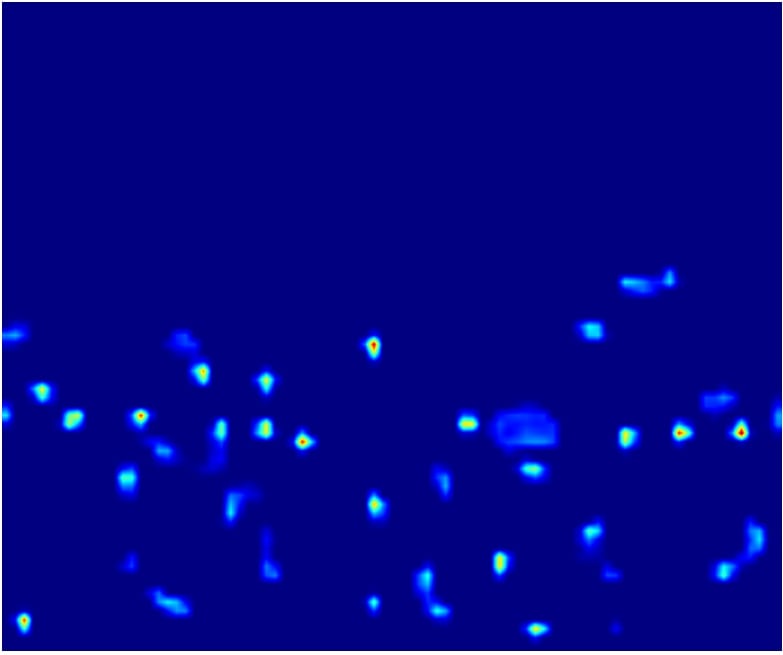} \hfill
        \includegraphics[width=0.195\textwidth]{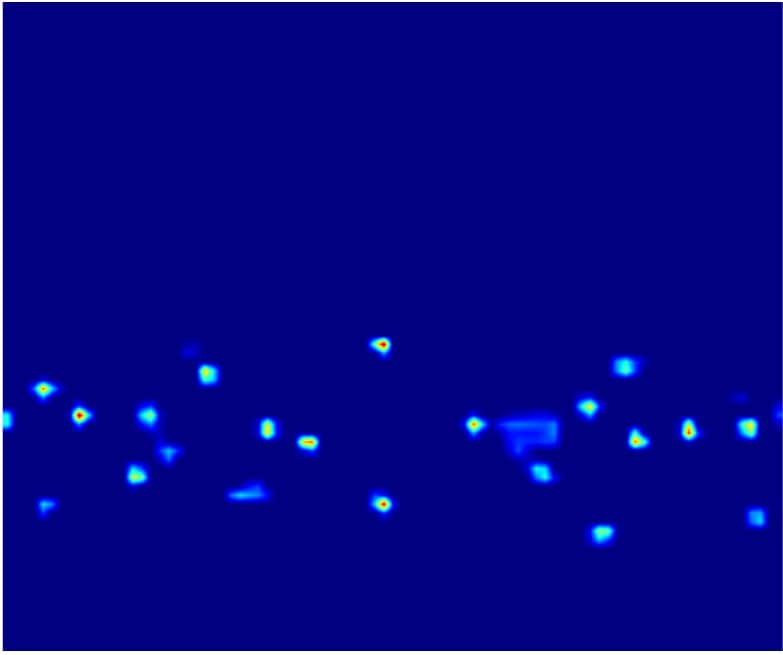}
    \end{minipage}
    \\
    \begin{minipage}[c]{0.02\textwidth}
    \rotatebox[origin = c]{90}{}
    \end{minipage}    
    \begin{minipage}[c]{0.975\textwidth}
    \centering
        \includegraphics[width=0.195\textwidth]{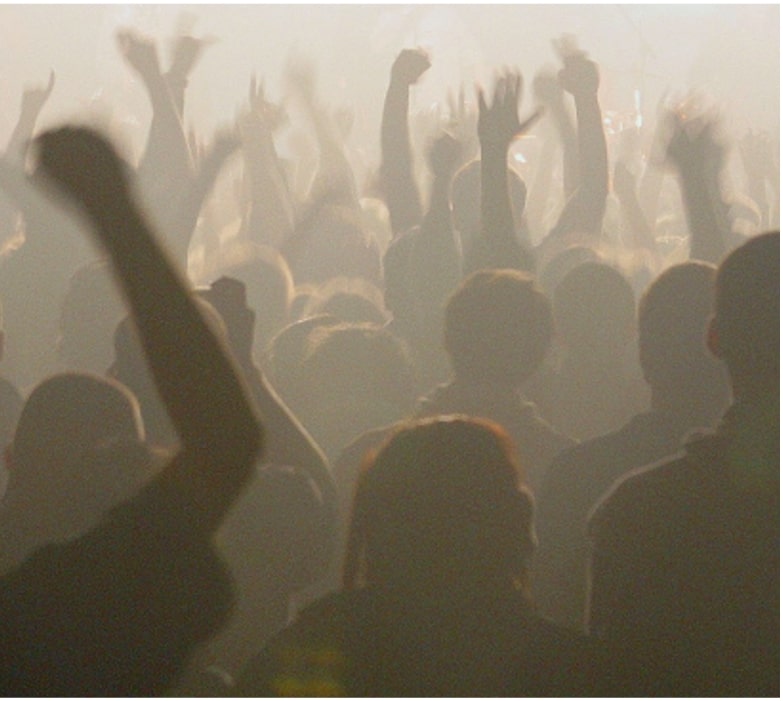} \hfill
        \includegraphics[width=0.195\textwidth]{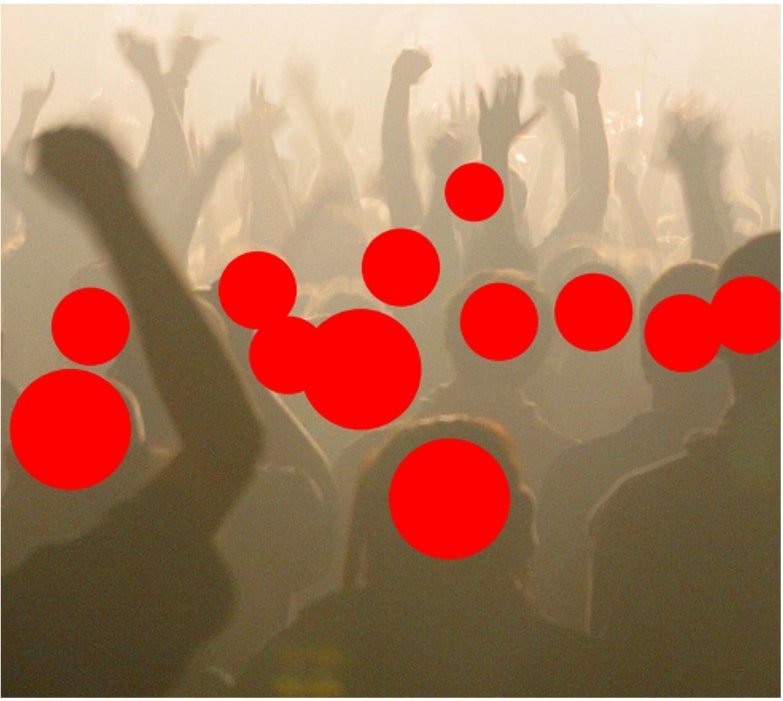} \hfill
        \includegraphics[width=0.195\textwidth]{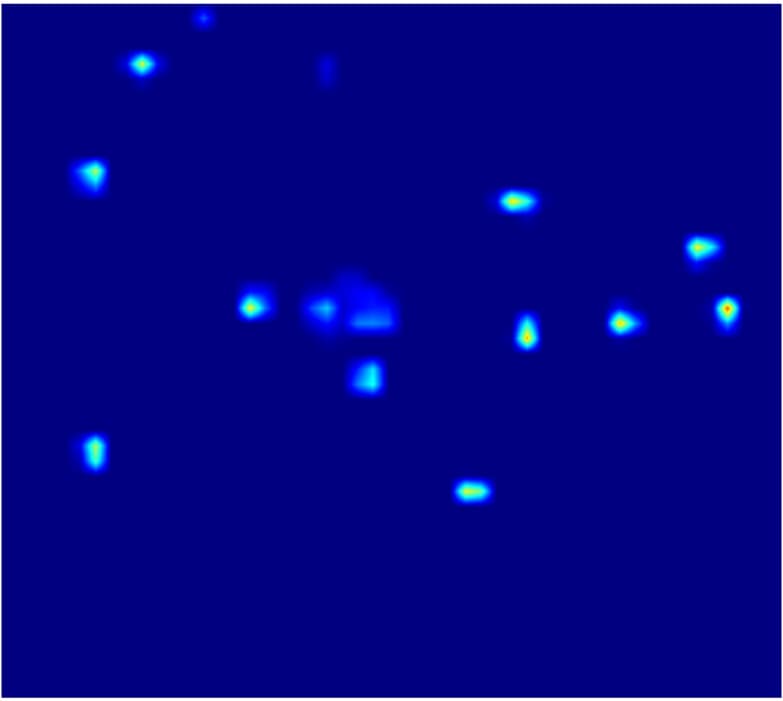} \hfill
        \includegraphics[width=0.195\textwidth]{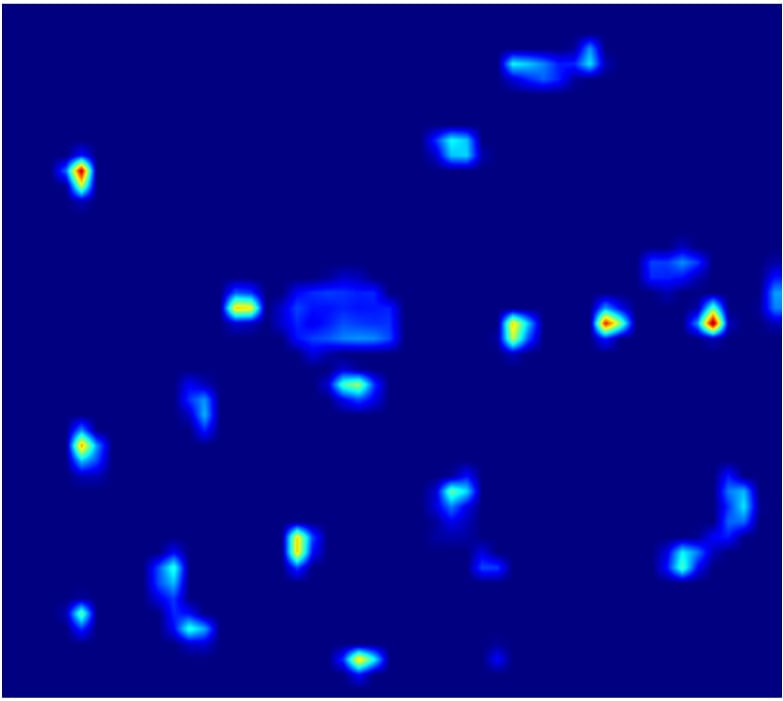} \hfill
        \includegraphics[width=0.195\textwidth]{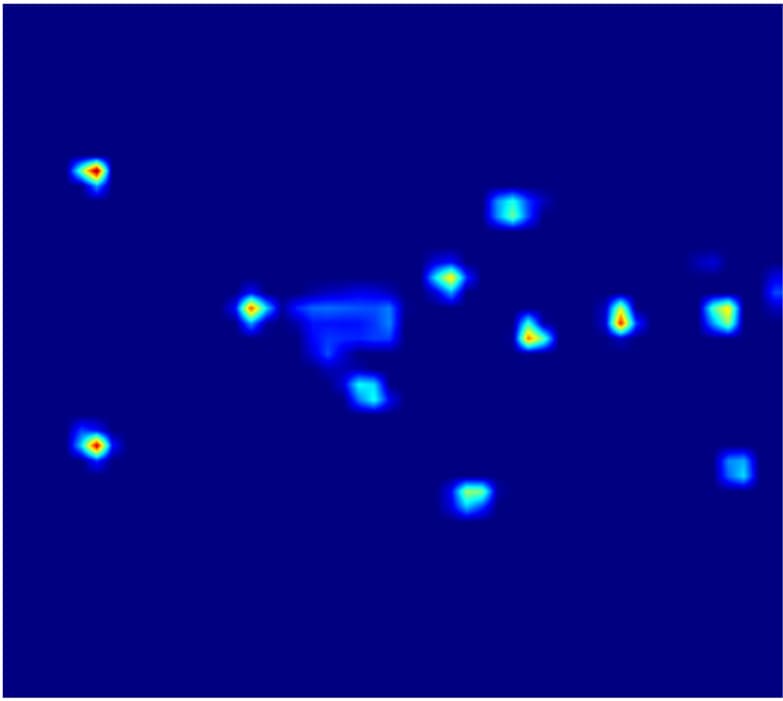}
    \end{minipage}
    \\
    \hspace{0.006\textwidth}
     \begin{minipage}[c]{0.195\textwidth}
      \centering    
        \text{}
    \end{minipage}
    \begin{minipage}[c]{0.195\textwidth}
     \centering
        \text{Count=22}
    \end{minipage}
         \begin{minipage}[c]{0.195\textwidth}
     \centering
        \text{Count=27}
    \end{minipage}
         \begin{minipage}[c]{0.195\textwidth}
     \centering
        \text{Count=35}
    \end{minipage}
         \begin{minipage}[c]{0.195\textwidth}
     \centering
        \text{Count=22}
    \end{minipage}
    \\
    \begin{minipage}[c]{0.02\textwidth}
    \rotatebox[origin = c]{90}{Snow}
    \end{minipage}
    \begin{minipage}[c]{0.975\textwidth}
    \centering
        \includegraphics[width=0.195\textwidth]{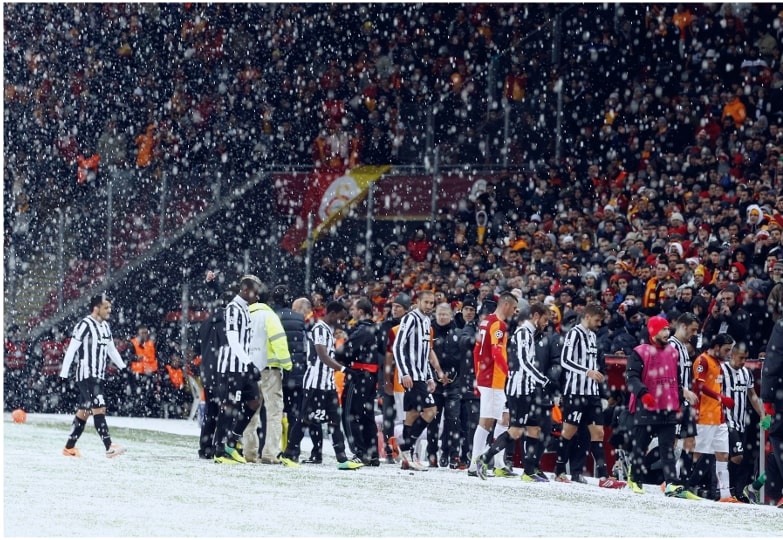} \hfill
        \includegraphics[width=0.195\textwidth]{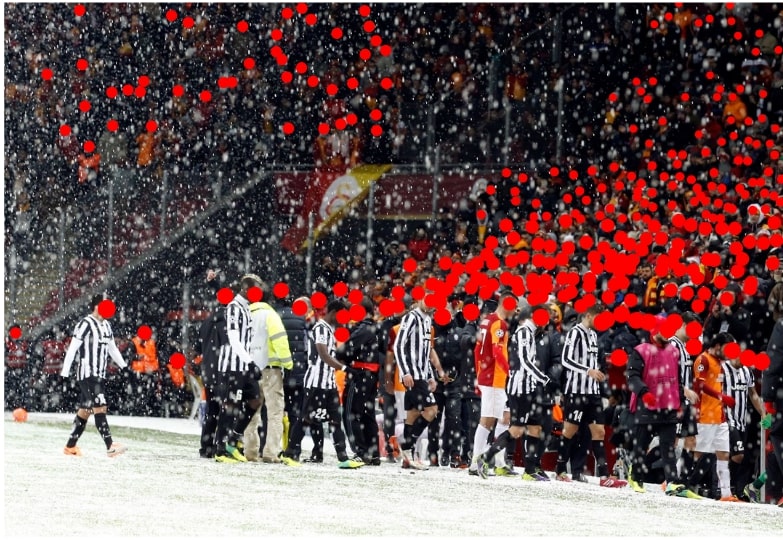} \hfill
        \includegraphics[width=0.195\textwidth]{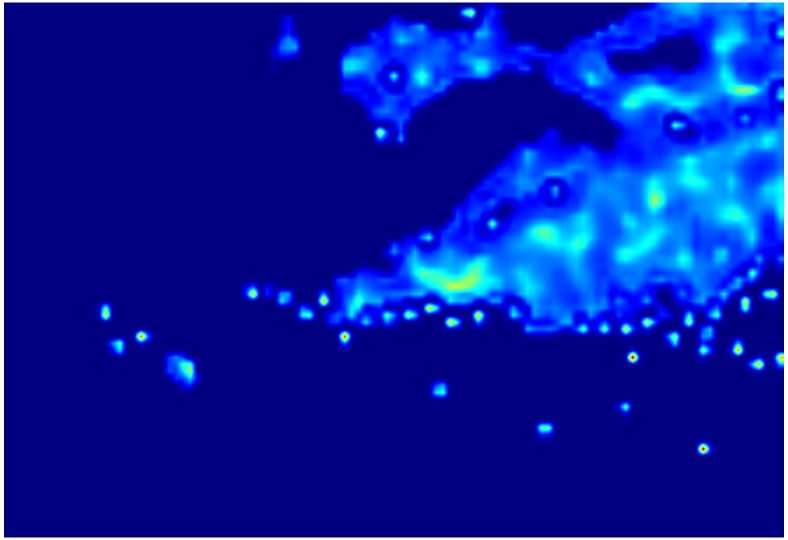} \hfill
        \includegraphics[width=0.195\textwidth]{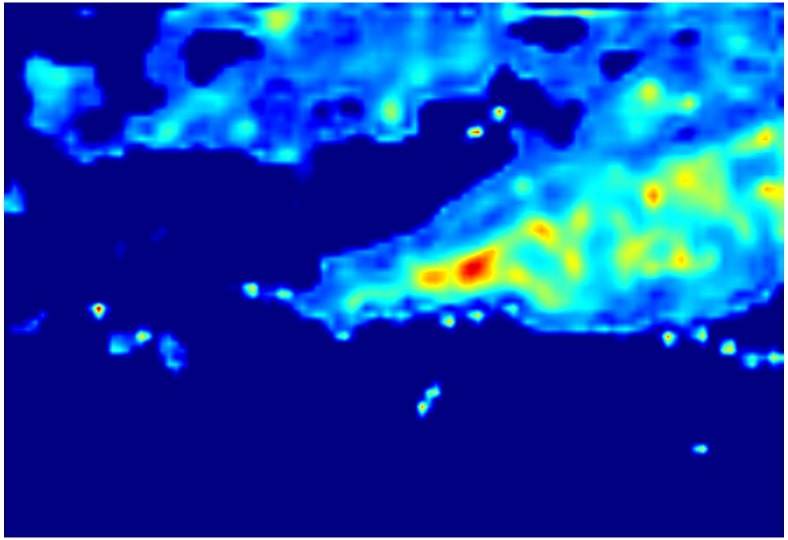} \hfill
        \includegraphics[width=0.195\textwidth]{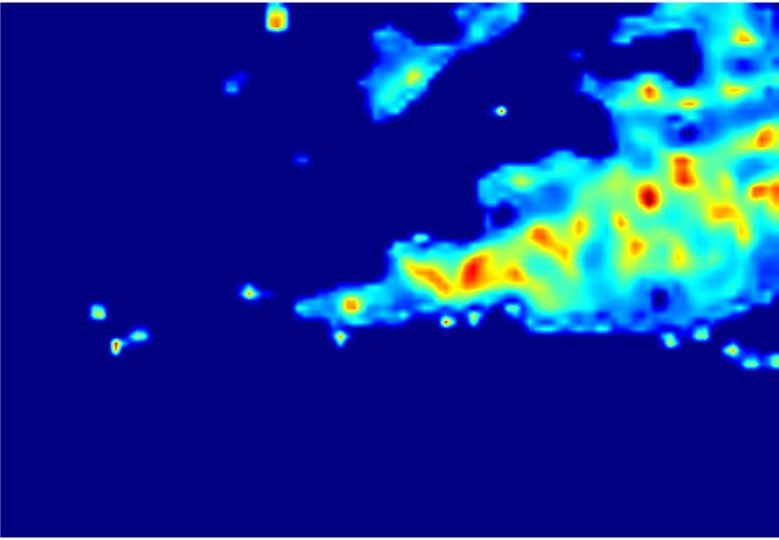}
    \end{minipage}
    \\
    \begin{minipage}[c]{0.02\textwidth}
    \rotatebox[origin = c]{90}{}
    \end{minipage}    
    \begin{minipage}[c]{0.975\textwidth}
    \centering
        \includegraphics[width=0.195\textwidth]{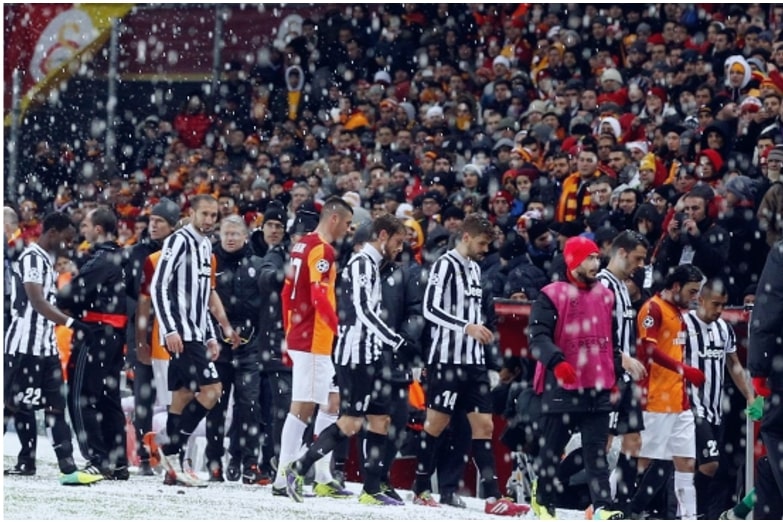} \hfill
        \includegraphics[width=0.195\textwidth]{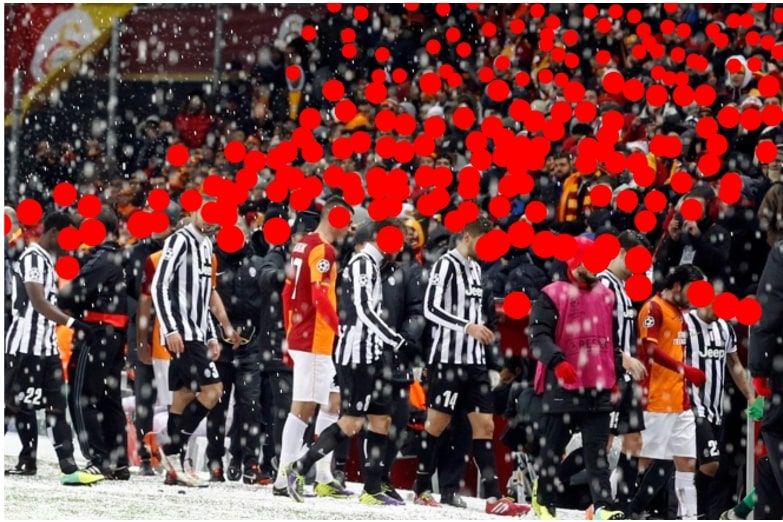} \hfill
        \includegraphics[width=0.195\textwidth]{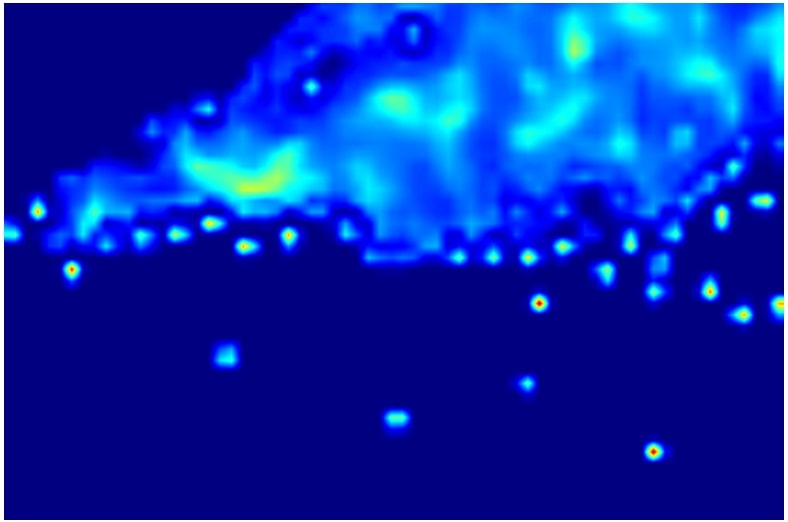} \hfill
        \includegraphics[width=0.195\textwidth]{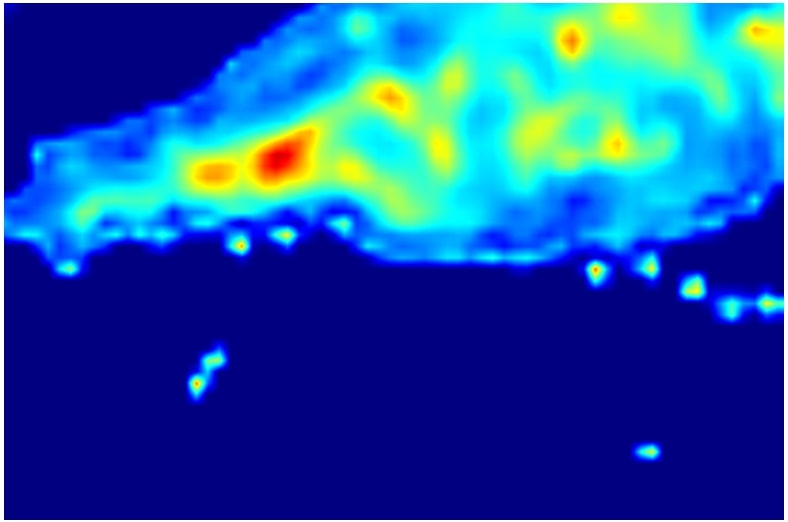} \hfill
        \includegraphics[width=0.195\textwidth]{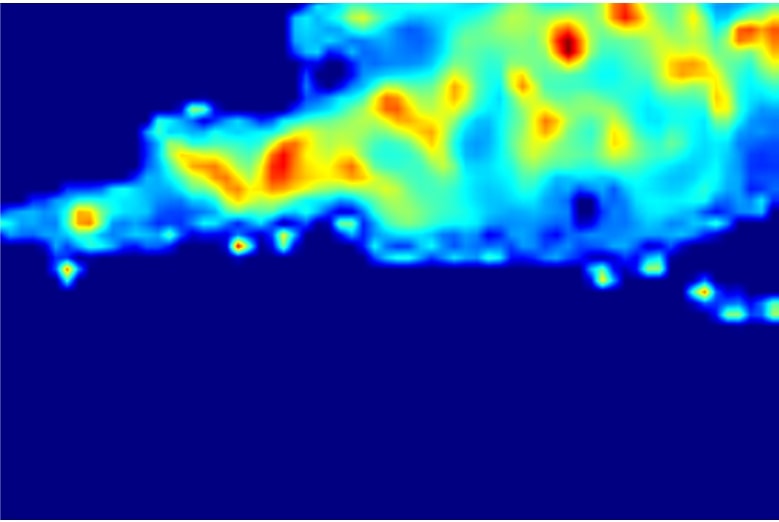}
    \end{minipage}
    \\
    \hspace{0.006\textwidth}
     \begin{minipage}[c]{0.195\textwidth}
      \centering    
        \text{}
    \end{minipage}
    \begin{minipage}[c]{0.195\textwidth}
     \centering
        \text{Count=285}
    \end{minipage}
         \begin{minipage}[c]{0.195\textwidth}
     \centering
        \text{Count=400}
    \end{minipage}
         \begin{minipage}[c]{0.195\textwidth}
     \centering
        \text{Count=473}
    \end{minipage}
         \begin{minipage}[c]{0.195\textwidth}
     \centering
        \text{Count=335}
    \end{minipage}
    \\
    \begin{minipage}[c]{0.02\textwidth}
    \rotatebox[origin = c]{90}{Rain}
    \end{minipage}
    \begin{minipage}[c]{0.975\textwidth}
    \centering
        \includegraphics[width=0.195\textwidth]{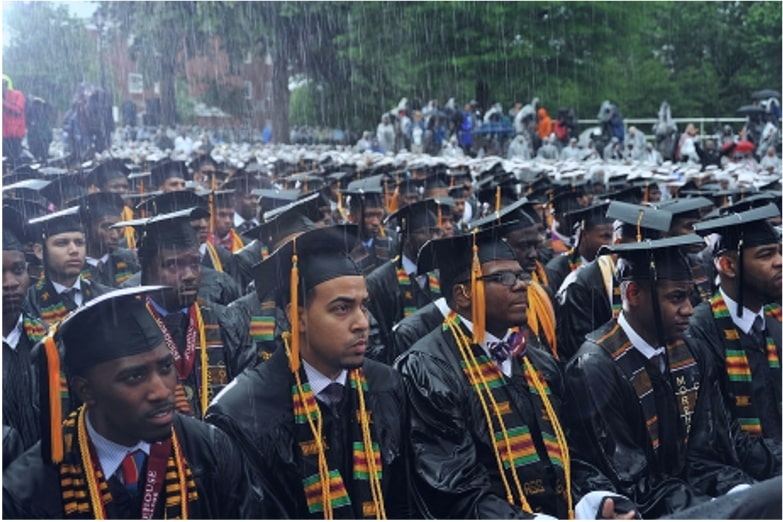} \hfill
        \includegraphics[width=0.195\textwidth]{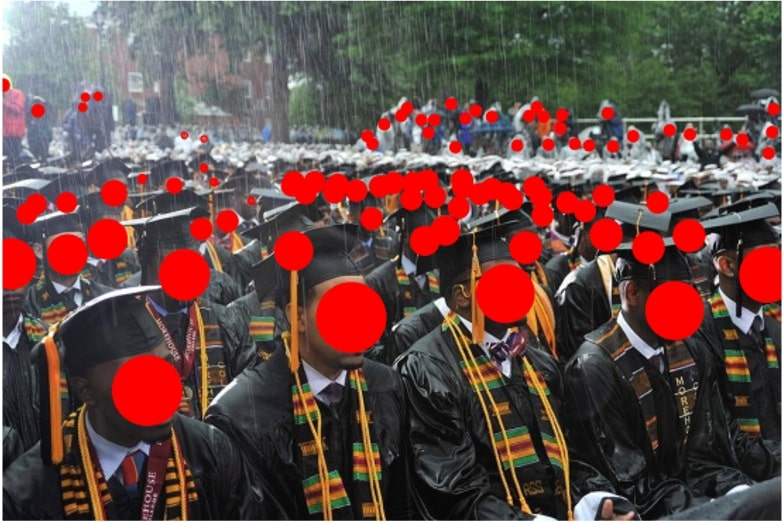} \hfill
        \includegraphics[width=0.195\textwidth]{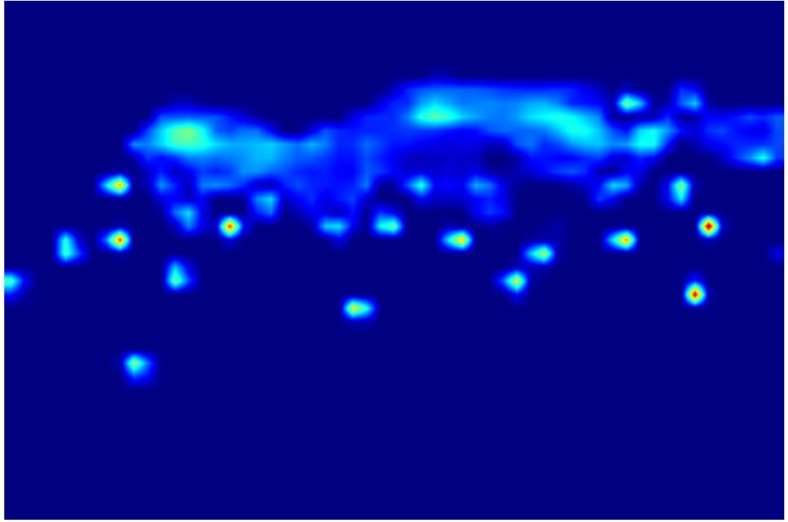} \hfill
        \includegraphics[width=0.195\textwidth]{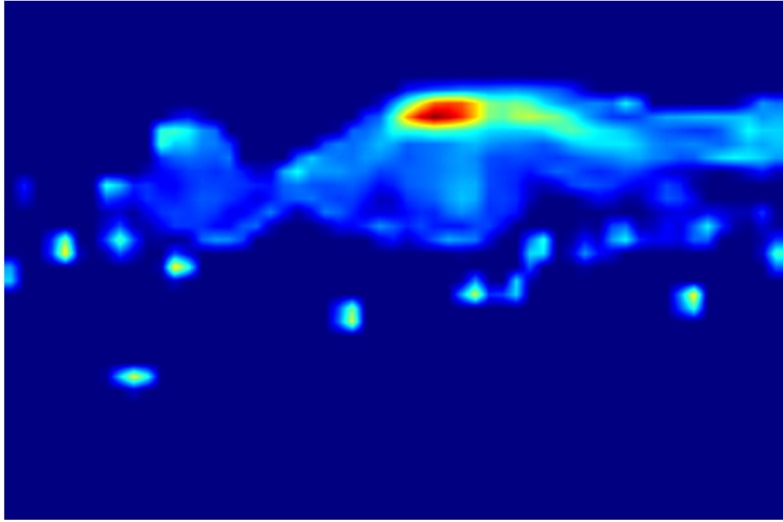} \hfill
        \includegraphics[width=0.195\textwidth]{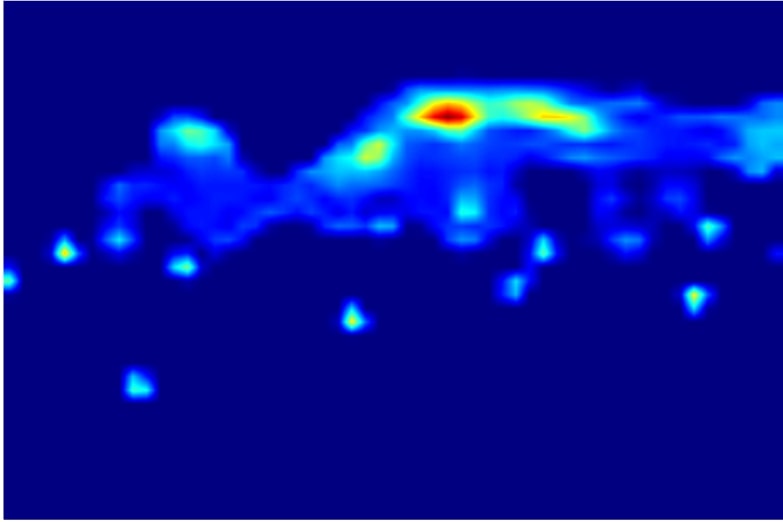}
    \end{minipage}
    \\
    \begin{minipage}[c]{0.02\textwidth}
    \rotatebox[origin = c]{90}{}
    \end{minipage}    
    \begin{minipage}[c]{0.975\textwidth}
    \centering
        \includegraphics[width=0.195\textwidth]{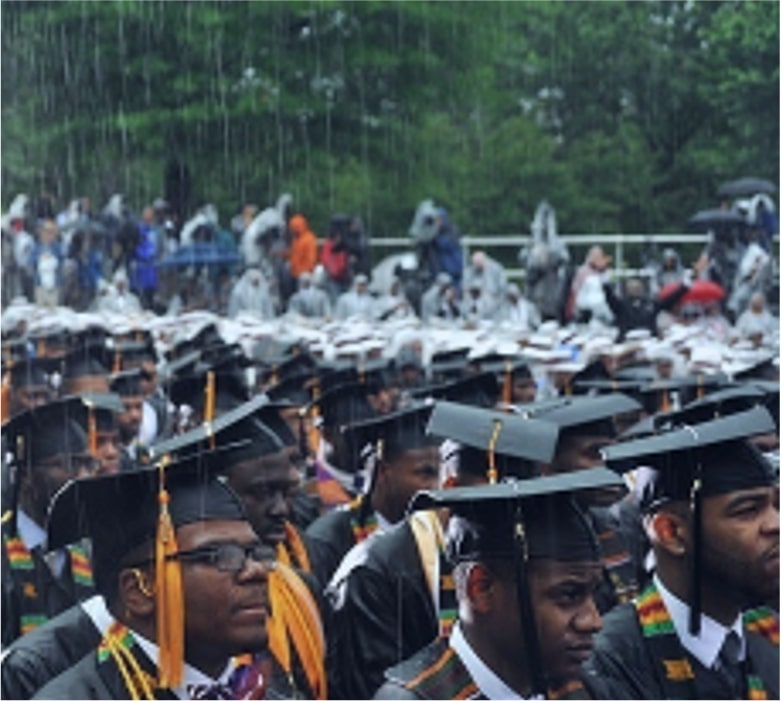} \hfill
        \includegraphics[width=0.195\textwidth]{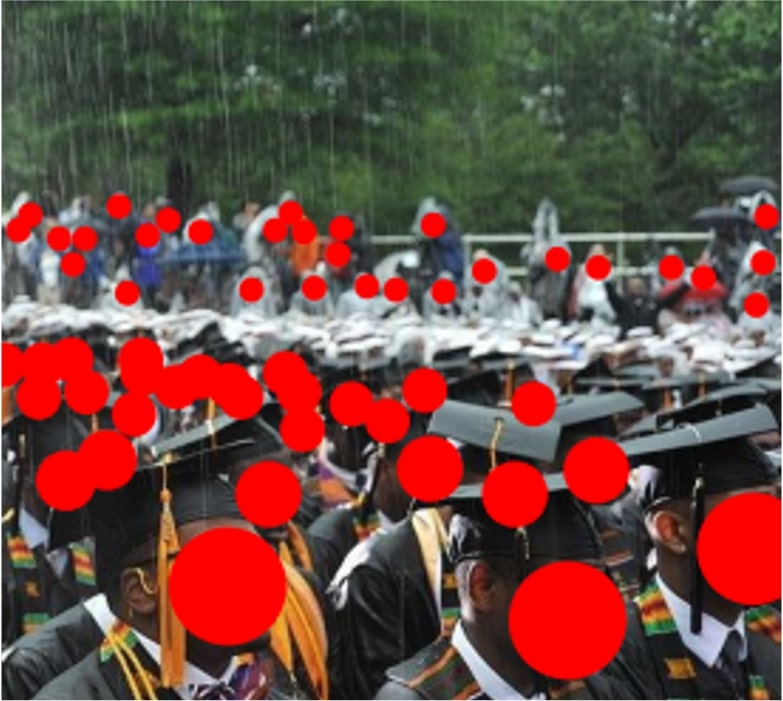} \hfill
        \includegraphics[width=0.195\textwidth]{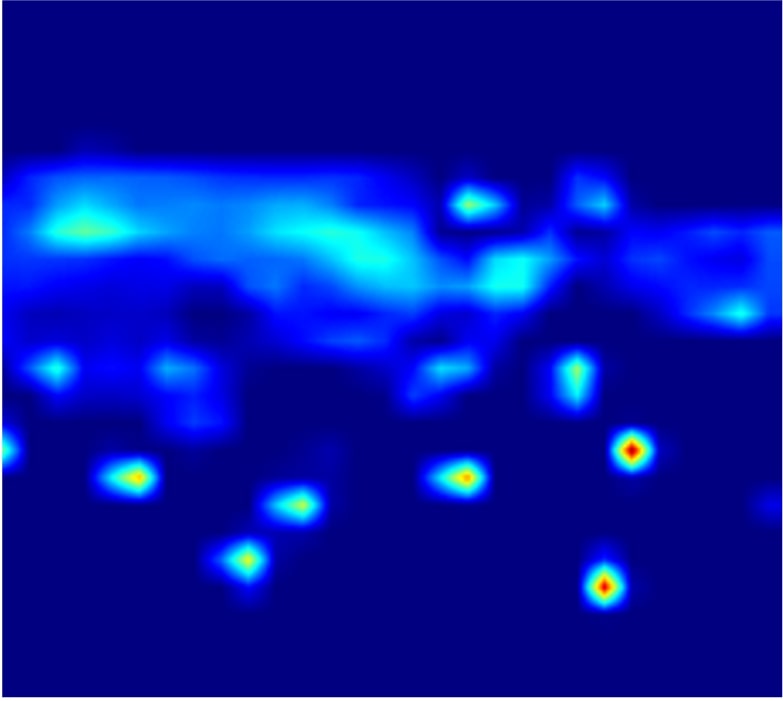} \hfill
        \includegraphics[width=0.195\textwidth]{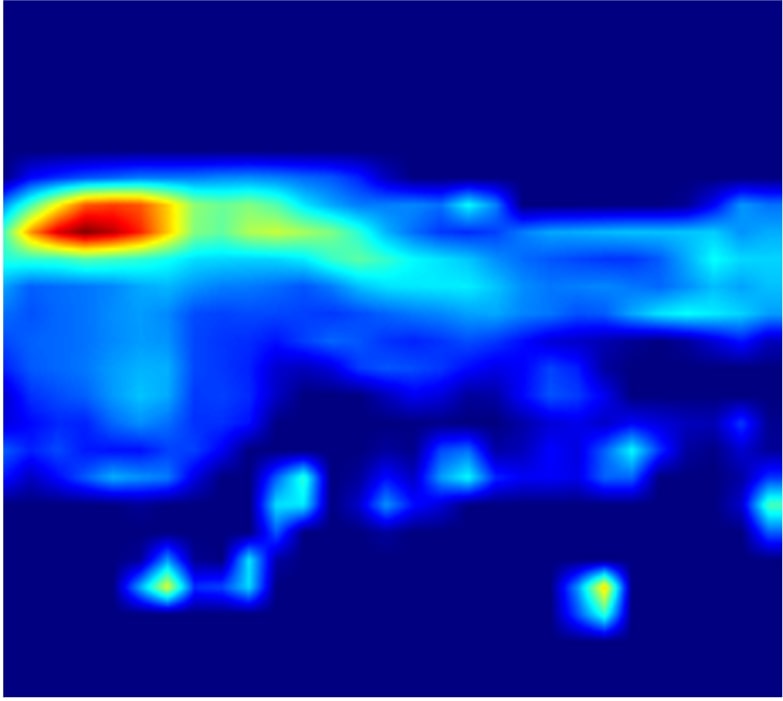} \hfill
        \includegraphics[width=0.195\textwidth]{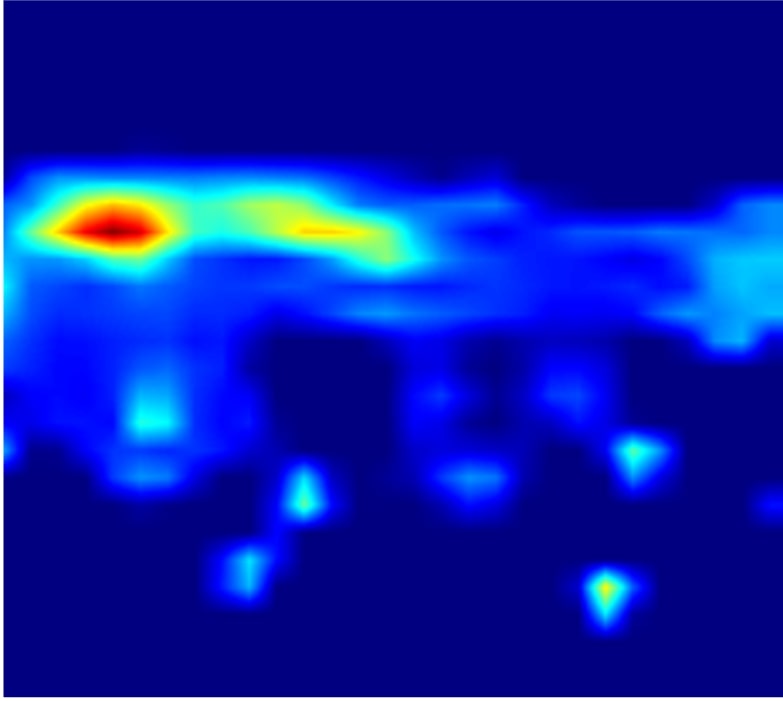}
    \end{minipage}
    \\
    \hspace{0.006\textwidth}
     \begin{minipage}[c]{0.195\textwidth}
      \centering    
        \text{}
    \end{minipage}
    \begin{minipage}[c]{0.195\textwidth}
     \centering
        \text{Count=94}
    \end{minipage}
         \begin{minipage}[c]{0.195\textwidth}
     \centering
        \text{Count=69}
    \end{minipage}
         \begin{minipage}[c]{0.195\textwidth}
     \centering
        \text{Count=75}
    \end{minipage}
         \begin{minipage}[c]{0.195\textwidth}
     \centering
        \text{Count=88}
    \end{minipage}
    \\
    \begin{minipage}[c]{0.02\textwidth}
    \rotatebox[origin = c]{90}{Clear}
    \end{minipage}
    \begin{minipage}[c]{0.975\textwidth}
    \centering
        \includegraphics[width=0.195\textwidth]{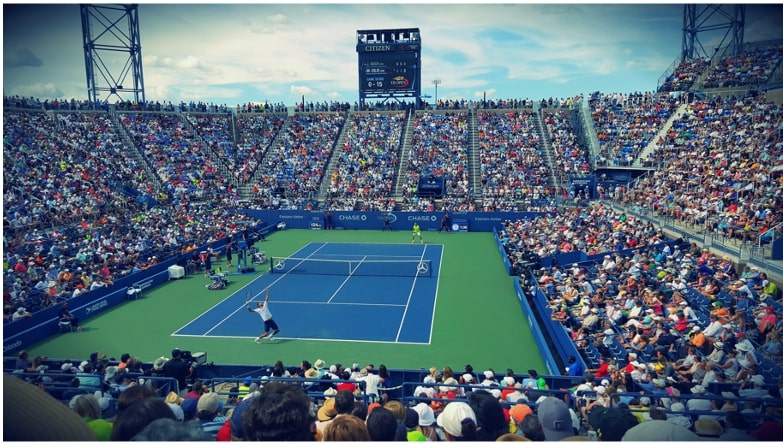} \hfill
        \includegraphics[width=0.195\textwidth]{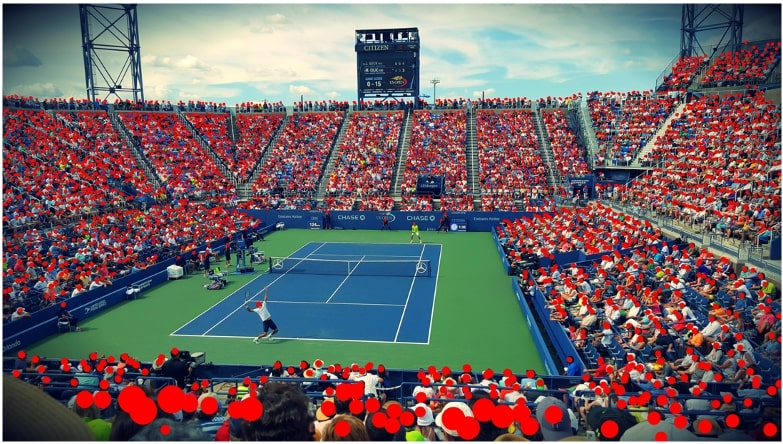} \hfill
        \includegraphics[width=0.195\textwidth]{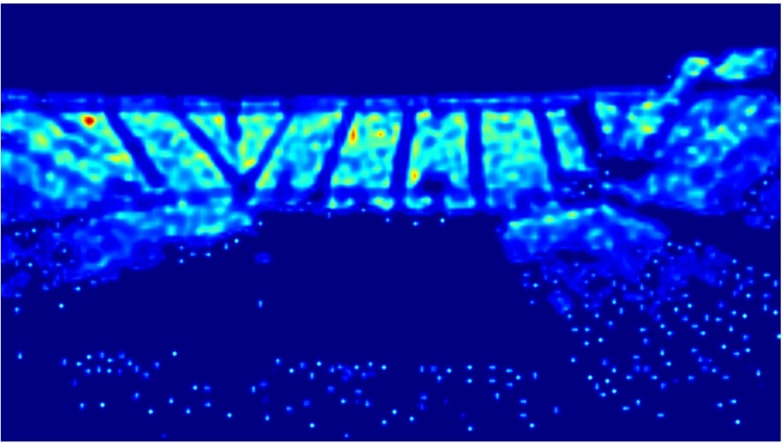} \hfill
        \includegraphics[width=0.195\textwidth]{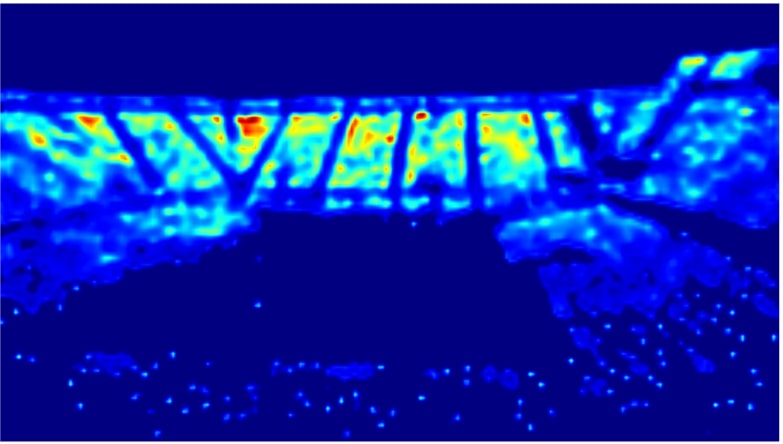} \hfill
        \includegraphics[width=0.195\textwidth]{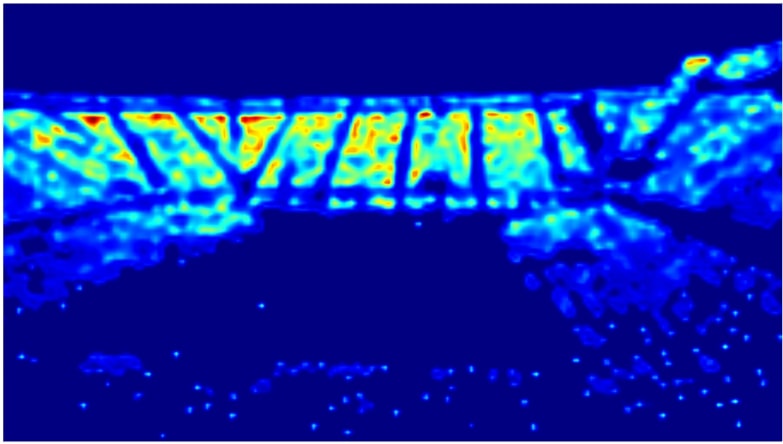}
    \end{minipage}
    \\
    \begin{minipage}[c]{0.02\textwidth}
    \rotatebox[origin = c]{90}{}
    \end{minipage}    
    \begin{minipage}[c]{0.975\textwidth}
    \centering
        \includegraphics[width=0.195\textwidth]{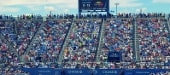} \hfill
        \includegraphics[width=0.195\textwidth]{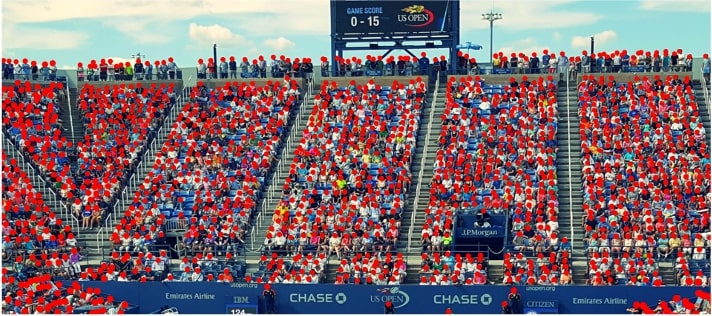} \hfill
        \includegraphics[width=0.195\textwidth]{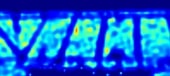} \hfill
        \includegraphics[width=0.195\textwidth]{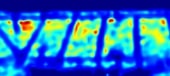} \hfill
        \includegraphics[width=0.195\textwidth]{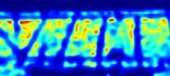}
    \end{minipage}
    \\
    \hspace{0.006\textwidth}
     \begin{minipage}[c]{0.195\textwidth}
      \centering    
        \text{}
    \end{minipage}
    \begin{minipage}[c]{0.195\textwidth}
     \centering
        \text{Count=3600}
    \end{minipage}
         \begin{minipage}[c]{0.195\textwidth}
     \centering
        \text{Count=3121}
    \end{minipage}
         \begin{minipage}[c]{0.195\textwidth}
     \centering
        \text{Count=2996}
    \end{minipage}
         \begin{minipage}[c]{0.195\textwidth}
     \centering
        \text{Count=3218}
    \end{minipage}
    
    \figmargin
\end{center}
    \caption{
        \textbf{Comparison of density maps of the proposed method and other methods in the adverse weather (i.e., haze, snow, rain) and clear scene.} The proposed method can compute more accurate density maps compared to the results estimated by other strategies.
    } 
    \label{fig:supp_density}
\end{figure*}

}

\clearpage

{\small
\twocolumn
\bibliographystyle{ieee_fullname}
\bibliography{egbib}
}

\end{document}